%% file: main.tex
\documentclass[10pt,twocolumn,letterpaper]{article}
\usepackage[pagenumbers]{cvpr}
\usepackage[accsupp]{axessibility}
\usepackage{graphicx}
\usepackage{amsmath}
\usepackage{amssymb}
\usepackage{booktabs}
\usepackage{cite}
\usepackage{multirow}
\usepackage{colortbl}
\usepackage[linesnumbered,lined,vlined,ruled,commentsnumbered,noend]{algorithm2e}
\usepackage[pagebackref=true, breaklinks=true, colorlinks,
 citecolor=citecolor, linkcolor=linkcolor, bookmarks=false]{hyperref}
\usepackage[capitalize]{cleveref}
\usepackage{listings}
\usepackage{listing}
\usepackage{xcolor}

\definecolor{citecolor}{HTML}{0071BC}
\definecolor{linkcolor}{HTML}{ED1C24}

\crefname{section}{Sec.}{Secs.}
\Crefname{section}{Section}{Sections}
\Crefname{table}{Table}{Tables}
\crefname{table}{Tab.}{Tabs.}
\Crefname{equation}{Equation}{Equations}
\crefname{equation}{eq.}{eqs.}

\input{math_commands}

\lstdefinestyle{mystyle}{
    backgroundcolor=\color{backcolour},   
    commentstyle=\color{codegreen},
    keywordstyle=\color{magenta},
    numberstyle=\tiny\color{codegray},
    stringstyle=\color{codepurple},
    basicstyle=\ttfamily\footnotesize,
    breakatwhitespace=false,         
    breaklines=true,                 
    captionpos=b,                    
    keepspaces=true,                 
    numbers=left,                    
    numbersep=5pt,                  
    showspaces=false,                
    showstringspaces=false,
    showtabs=false,                  
    tabsize=2
}

\definecolor{codegreen}{rgb}{0,0.6,0}
\definecolor{codegray}{rgb}{0.5,0.5,0.5}
\definecolor{codepurple}{rgb}{0.58,0,0.82}
\definecolor{backcolour}{rgb}{0.95,0.95,0.92}

\makeatletter
\renewcommand{\paragraph}{%
  \@startsection{paragraph}{4}%
  {\z@}{0.25em}{-1em}%
  {\normalfont\normalsize\bfseries}%
}
\newcommand{\myparagraphsupp}[1]{\vspace{0.1em}\noindent{\textcolor[rgb]{0.93,0.47,0.32}{#1}}}
\newcommand{\mycaptionsupp}[1]{{\textcolor[rgb]{0.93,0.47,0.32}{#1}}}
\makeatother

\definecolor{mygray}{gray}{0.6}
\definecolor{mygray-bg}{gray}{0.9}

\newcommand{\beginsupp}{%
        \setcounter{table}{0}
        \renewcommand{\thetable}{S\arabic{table}}%
        \setcounter{figure}{0}
        \renewcommand{\thefigure}{S\arabic{figure}}%
     }

\begin{document}
\title{Continual Detection Transformer for Incremental Object Detection}
\author{Yaoyao Liu$^{1}$ 
\quad Bernt Schiele$^{1}$ 
\quad Andrea Vedaldi$^{2}$
\quad Christian Rupprecht$^{2}$\\
\\
\small $^{1}$Max Planck Institute for Informatics, Saarland Informatics Campus\\
\small $^{2}$Visual Geometry Group, Department of Engineering Science, University of Oxford\\
\small {
 \texttt{\{yaoyao.liu, schiele\}@mpi-inf.mpg.de}}  \quad  {\texttt{\{vedaldi, chrisr\}@robots.ox.ac.uk}}
}
\maketitle

\input{sections/0_abstract}
\input{sections/1_introduction}
\input{sections/2_related_work}
\input{sections/3_method}
\input{sections/4_experiments}

\input{sections/5_conclusion}

{\small\bibliographystyle{ieee_fullname}\bibliography{references}}

\clearpage
\input{supplementary/supplementary}
\end{document}

%% file: math_commands.tex
\usepackage{amsmath,amsfonts,bm}

\def\eqref#1{equation~\ref{#1}}

\def\1{\bm{1}}

\DeclareMathAlphabet{\mathsfit}{\encodingdefault}{\sfdefault}{m}{sl}
\SetMathAlphabet{\mathsfit}{bold}{\encodingdefault}{\sfdefault}{bx}{n}



%% file: sections/0_abstract.tex
\begin{abstract}
Incremental object detection (IOD) aims to train an object detector in phases, each with annotations for new object categories. As other incremental settings, IOD is subject to catastrophic forgetting, which is often addressed by techniques such as knowledge distillation (KD) and exemplar replay (ER). However, KD and ER do not work well if applied directly to state-of-the-art transformer-based object detectors such as Deformable DETR~\cite{Zhu2021DeformableDETR} and UP-DETR~\cite{Dai2021UPDETR}. In this paper, we solve these issues by proposing a ContinuaL DEtection TRansformer (CL-DETR), a new method for transformer-based IOD which enables effective usage of KD and ER in this context. First, we introduce a Detector Knowledge Distillation (DKD) loss, focusing on the most informative and reliable predictions from old versions of the model, ignoring redundant background predictions, and ensuring compatibility with the available ground-truth labels. We also improve ER by proposing a calibration strategy to preserve the label distribution of the training set, therefore better matching training and testing statistics. We conduct extensive experiments on COCO 2017 and demonstrate that CL-DETR achieves state-of-the-art results in the IOD setting.\footnote{Code: \href{https://lyy.mpi-inf.mpg.de/CL-DETR/}{https://lyy.mpi-inf.mpg.de/CL-DETR/}}
\end{abstract}

%% file: sections/1_introduction.tex
\section{Introduction}%
\label{sec_intro}

Humans inherently learn in an incremental manner, acquiring new concepts over time without forgetting previous ones.
In contrast, machine learning suffers from \emph{catastrophic forgetting}~\cite{McCloskey1989Catastrophic,McRae1993Catastrophic,Kirkpatrick2017Overcoming}, where learning from non-i.i.d.~data can override knowledge acquired previously.
Unsurprisingly, forgetting also affects object detection~\cite{Feng2022ElasticResponse,Aljundi2019TaskFree,Peng2020FasterILOD,Shmelkov2017Incremental,Verwimp2022ReExamining,Yang2022ContinualOD,KJ2021IODMeta}.
In this context, the problem was formalized by Shmelkov~\etal.~\cite{Shmelkov2017Incremental}, who defined an incremental object detection~(IOD) protocol, where the training samples for different object categories are observed in phases, restricting the ability of the trainer to access past data.

Popular methods to address forgetting in tasks other than detection include Knowledge Distillation (KD) and Exemplar Replay (ER).
KD~\cite{LiH2018LwF,Douillard2020PODNet,Hou2019LUCIR,Hu2021CausalEffect,Zhao2020Maintaining} uses regularization in an attempt to preserve previous knowledge when training the model on new data.
The key idea is to encourage the new model's logits or feature maps to be close to those of the old model.
ER methods~\cite{Rebuffi2017iCaRL,Liu2020Mnemonics,Liu2021RMM,Wang2022Memory,Castro18EndToEnd,Liu2020Generative} work instead by memorising some of the past training data (the \emph{exemplars}), replaying them in the following phases to ``remember'' the old object categories.

\input{figures/teaser}

Recent state-of-the-art results in object detection have been achieved by a family of transformer-based architectures that include DETR~\cite{Carion2020DETR}, Deformable DETR~\cite{Zhu2021DeformableDETR} and UP-DETR~\cite{Dai2021UPDETR}.
In this paper, we show that KD and ER do not work well if applied directly to these models.
For instance, in \cref{fig_teaser} we show that applying KD and ER to Deformable DETR leads to much worse results compared to training with all data accessible in each phase (\ie, the standard non-incremental setting).

We identify two main issues that cause this drop in performance.
First, transformer-based detectors work by testing a large number of object hypotheses in parallel.
Because the number of hypotheses is much larger than the typical number of objects in an image, most of them are negative, resulting in an unbalanced KD loss.
Furthermore, because both old and new object categories can co-exist in any given training image, the KD loss and regular training objective can provide contradictory evidence.
Second, ER methods for image classification try to sample the same number of exemplars for each category.
In IOD, this is not a good strategy because the true object category distribution is typically highly skewed.
Balanced sampling causes a mismatch between the training and testing data statistics.

In this paper, we solve these issues by proposing \emph{ContinuaL DEtection TRansformer} (CL-DETR), a new method for transformer-based IOD which enables effective usage of KD and ER in this context.
CL-DETR introduces the concept of \emph{Detector Knowledge Distillation} (DKD), selecting the most confident object predictions from the old model, merging them with the ground-truth labels for the new categories while resolving conflicts, and applying standard joint bipartite matching between the merged labels and the current model's predictions for training.
This approach subsumes the KD loss, applying it only for foreground predictions correctly matched to the appropriate model's hypotheses.
CL-DETR also improves ER by introducing a new calibration strategy to preserve the distribution of object categories observed in the training data.
This is obtained by carefully engineering the set of exemplars remembered to match the desired distribution.
Furthermore, each phase consists of a main training step followed by a smaller one focusing on better calibrating the model.

We also propose a more realistic variant of the IOD benchmark protocol.
In previous works~\cite{Shmelkov2017Incremental,Feng2022ElasticResponse}, in each phase, the incremental detector is allowed to observe all images that contain a certain type of object.
Because images often contain a mix of object classes, both old and new, this means that the same images can be observed in different training phases.
This is incompatible with the standard definition of incremental learning~\cite{Rebuffi2017iCaRL,Liu2020Mnemonics,Hou2019LUCIR} where, with the exception of the examples deliberately stored in the exemplar memory, the images observed in different phases do not repeat.
We redefine the IOD protocol to avoid this issue.

We demonstrate CL-DETR by applying it to different transformer-based detectors including Deformable DETR~\cite{Zhu2021DeformableDETR} and UP-DETR~\cite{Dai2021UPDETR}.
As shown in~\cref{fig_teaser}, our results on COCO 2017 show that CL-DETR leads to significant improvements compared to the baseline, boosting AP by $4.2$ percentage points compared to a direct application of KD and ER to the underlying detector model.
We further study and justify our modelling choices via ablations.

To summarise, we make \textbf{four contributions}:
(1) The DKD loss that improves KD for knowledge distillation by resolving conflicts between distilled knowledge and new evidence and by ignoring redundant background detections;
(2) A calibration strategy for ER to match the stored exemplars to the training set distribution;
(3) A revised IOD benchmark protocol that avoids observing the same images in different training phases;
(4) Extensive experiments on COCO 2017, including state-of-the-art results, an in-depth ablation study, and further visualizations.

%% file: figures/teaser.tex
\begin{figure}
\centering
\includegraphics[width=0.43\textwidth]{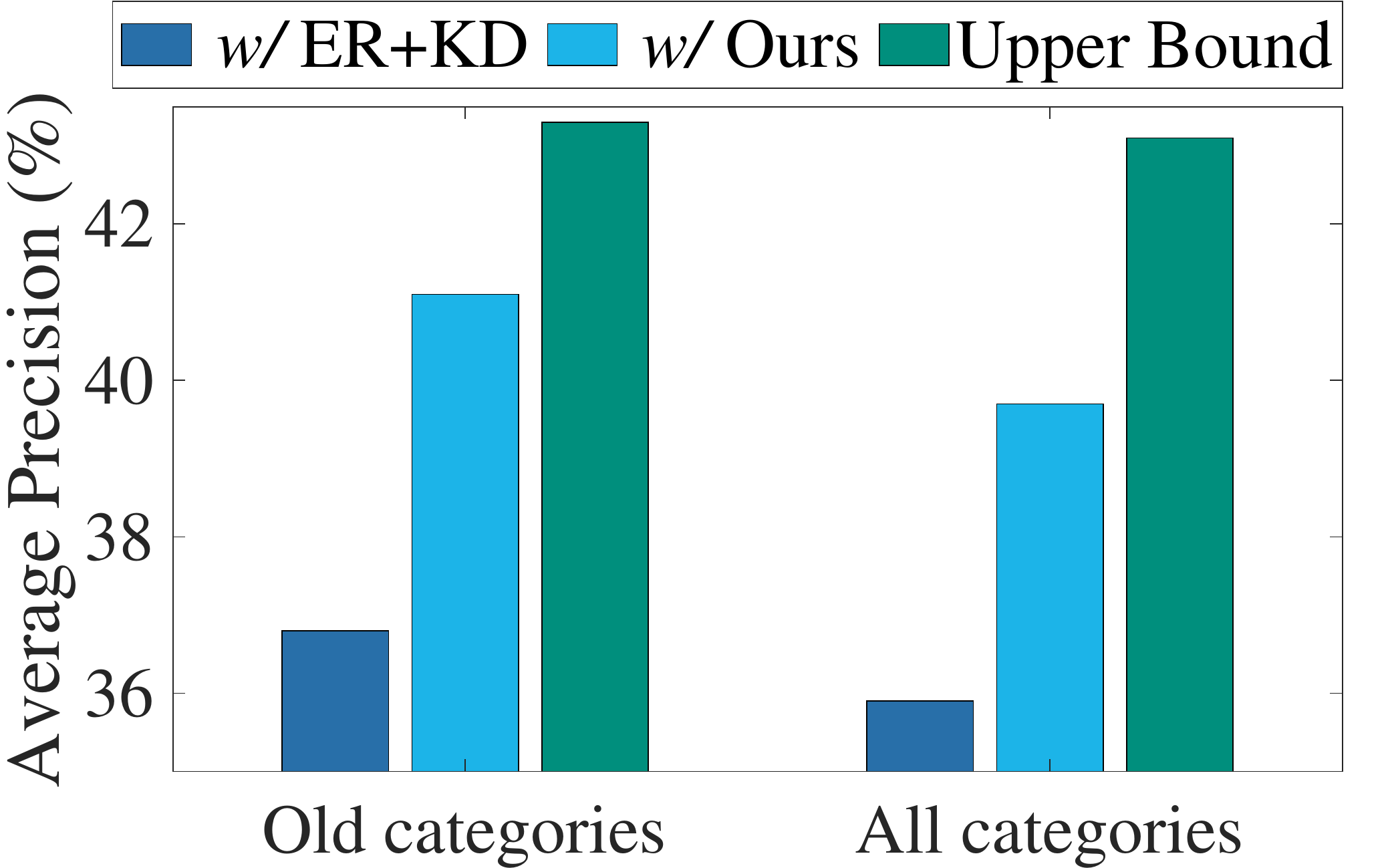}
\vspace{-0.1cm}
\caption{The final Average Precision (AP, $\%$) of two-phase incremental object detection on COCO 2017. We observe $70$ and $10$ categories in the first and second phases, respectively. The baseline is Deformable DETR~\cite{Zhu2021DeformableDETR}. ``Upper bound'' shows the results of joint training with all previous data accessible in each phase.}%
\label{fig_teaser}
\vspace{-0.2cm}
\end{figure}

%% file: sections/2_related_work.tex
\section{Related Work}
\label{sec_related_work}

\paragraph{Incremental learning.}

Incremental learning (also known as continual learning~\cite{De2021ContinualSurvey,Aljundi2019TaskFree,LopezPazR2017GEM} and lifelong learning~\cite{Aljundii2017ExpertGate,Chen2018Lifelong,Chaudhry2019AGEM}) aims at learning models in phases that focus on different subsets of the label space.
Recent incremental learning methods can be divided into two categories:
(i) Knowledge Distillation~(KD) tries to preserve the knowledge capture in a previous version of the model by matching logits~\cite{LiH2018LwF,Rebuffi2017iCaRL}, feature maps~\cite{Douillard2020PODNet}, or other information~\cite{Tao2020Topology,Wang2022FOSTER,Simon2021Learning,Joseph2022Energy,Pourkeshavarzi2022Looking,Liu2023Online} in the new model.
(ii) Exemplar Replay (ER) methods build a reservoir of samples or exemplars from old training rounds~\cite{Rebuffi2017iCaRL,Shin2017GenerativeReplay,Liu2020Mnemonics,Prabhu2020GDumb,Bang2021Rainbow} and replay them in successive training phases as a way of recalling past knowledge.
KD and ER are the starting point of our method.

\paragraph{Incremental object detection (IOD).}

IOD applies incremental learning to object detection specifically.
This is more challenging than incremental image classification, as images can contain multiple objects, both of old and new types, with only the new types being annotated in any given training phase.
Both KD and ER have been applied to detection before.
{}\cite{Shmelkov2017Incremental} applies KD to the output of Faster R-CNN~\cite{Girshick2015FastRCNN}.
Inspired by this, recent IOD methods extended the KD framework to other detectors (\eg, Faster-RCNN~\cite{Ren2017FasterRCNN} and GFL~\cite{Li2020GFL}) by adding KD terms on the intermediate feature maps~\cite{Yang2022Multi,Zhou2020LifelongOD,Feng2022ElasticResponse} and region proposal networks~\cite{Chen2019New,Hao2019EndtoEnd,Peng2020FasterILOD}.
{}\cite{Joseph2021TowardsOpenWorld} proposes instead to store a set of exemplars and fine-tune the model on the exemplars after each incremental step.
{}\cite{Liu2020MultiTask} proposes an adaptive sampling strategy to achieve more efficient exemplar selection for IOD\@.

However, existing IOD methods are designed based on conventional detectors such as Faster-RCNN~\cite{Ren2017FasterRCNN} and GFL~\cite{Li2020GFL}.
In this work, we show that a direct application of KD and ER to current state-of-the-art transformer-based detectors such as Deformable DETR~\cite{Zhu2021DeformableDETR} and UP-DETR~\cite{Dai2021UPDETR} does not work well and we propose fixes to this issue.

\paragraph{Transformer-based object detection.}

DEtection TRansformer (DETR)~\cite{Carion2020DETR} proposes an elegant architecture for object detection based on a visual transformer~\cite{Vaswani2017Transformer}.
Compared to pre-transformer approaches, DETR eliminates the need for non-maximum suppression in post-processing because self-attention can learn to remove duplicated detection by itself.
This is achieved  by using the Hungarian loss, matching each object hypothesis to exactly one target or background using bipartite matching~\cite{Sun2021Rethinking}. Deformable DETR~\cite{Zhu2021DeformableDETR} improves the performance of DETR, particularly for small objects, via  sparse attention on multi-level feature maps.
UP-DETR~\cite{Dai2021UPDETR} leverages unsupervised learning to pre-train the parameters of the encoder and decoder in DETR to further boost the performance.

Our method does not fundamentally change these detectors and is in fact applicable to all similar ones.
Instead, it proposes broadly-applicable changes that make transformer-based detectors work well in combination with KD and ER for the IOD problem.

%% file: sections/3_method.tex
\section{Methodology}%
\label{sec_method}

\newcommand{\D}{\mathcal{D}}
\newcommand{\N}{\mathcal{N}}
\newcommand{\C}{\mathcal{C}}
\newcommand{\F}{\mathcal{F}}
\newcommand{\PP}{\mathcal{P}}
\newcommand{\Q}{\mathcal{Q}}
\newcommand{\EE}{\mathcal{E}}

\input{figures/fig_framework_1}

After defining the incremental detection problem~(\cref{sec_preliminaries}) and providing the necessary background~(\cref{subsec_transformer_detector}), we introduce ContinuaL DEtection TRansformer (CL-DETR), a new method for incremental object detection that extends  DETR-like detectors with knowledge distillation (KD\@; \cref{subsec_mpdistillation}) and exemplar replay~(ER\@; \cref{subsec_learning_calibration}).

\subsection{Incremental object detection}%
\label{sec_preliminaries}

In \emph{incremental object detection} (IOD) the goal is to train a detector in phases, where in each phase the model is only given annotations for a subset of the object categories.
Formally, let $\D = \{(x,y)\}$ be a dataset of images $x$ with corresponding object annotations $y$, such as COCO~2017~\cite{Lin2014COCO}, and let $\C=\{1,\dots,C\}$ be the set of object categories.
We adapt such a dataset for benchmarking IOD as follows.
First, we partition $\D$ and $\C$ into $M$ subsets $\D = \D_1\cup\dots\cup\D_M$ and $\C=\C_1\cup\dots\cup\C_M$, one for each training phase.
For each phase $i$, we modify the samples $(x,y)\in\D_i$ so that $y$ only contains annotations for objects of class $\C_i$ and drop the others.\footnote{In this way, some images end up containing no annotated objects.}

In phase $i$ of training, the model is only allowed to observe images $\D_i$ with annotations for objects of types $\C_i \subset\C$.
Notably, images can and do contain objects of any possible type $\C$, but only types $\C_{i}$ are annotated in this phase.
After phase $i$ is complete, training switches to the next phase $i+1$, so the model observes different images $\D_{i+1}$ and annotations for objects of different types $\C_{i+1}$.

For exemplar replay, we relax this training protocol and allow the model to memorise a small number of \emph{exemplars} $\EE_i \subset \D_i$ from the previous phases.
In this case, the model is trained on the union $\D_i \cup \EE_{1:i-1}$ where
$
\EE_{1:i-1}
=
\EE_1 \cup \dots \cup \EE_{i-1}
$
forms the exemplar memory.

Note that this is a stricter and improved protocol compared to prior works in IOD~\cite{Shmelkov2017Incremental,Feng2022ElasticResponse}.
In these works, the model is still presented a subset of annotations restricted to classes $\C_i$ in each phase;
however, $\D_i\subset \D$ is defined as the subset of all images that contain objects of type $\C_i$.
Because images contain a mix of object categories that can span different subsets $\C_i$, this means that different subsets $\D_i$ can overlap, so that the same images can be observed multiple times in different phases.
This  violates the standard definition of incremental learning~\cite{Rebuffi2017iCaRL,Liu2020Mnemonics,Hou2019LUCIR} which assumes that different samples are observed in different phases.
Our setting retains this property.

\subsection{Transformer-based detectors}%
\label{subsec_transformer_detector}

State-of-the-art methods like DETR~\cite{Zhu2021DeformableDETR,Carion2020DETR,Dai2021UPDETR,Liu2021WBDETR,Sun2021Rethinking,Zhang2022DINO} build on powerful visual transformers to solve the object detection problem.
In order to motivate and explain our method, we first review briefly how they work.

With reference to \cref{fig_framework_1}, the model $\Phi$ takes as input an image
$
x \in \mathbb{R}^{3\times H\times W}
$
and outputs the object predictions $\hat y = \Phi(x)$ using a number of attention and self-attention layers.
The output
$
\hat y
= (\hat y_j)_{j\in\N}
$
is a sequence $\N=\{1,\dots,N\}$ of object predictions
$
\hat{y}_j = (\hat p_j, \hat b_j),
$
consisting of a class probability vector
$
\hat p_j : \C \cup \{ \phi \} \rightarrow [0,1]
$
and a vector $b_j \in [0,1]^4$ specifying the centre and size of the object bounding box relative to the image size.
Note that the support of $\hat p_j$ includes element $\phi$ that denotes the background class, or `no object' (hence, $\hat p_j$ has $C+1$ dimensions).

The object predictions correspond to a fixed set of \emph{object queries} internal to the model.
Each query is thus mapped to an object instance or background.
The order of the queries is conceptually immaterial, but queries are fixed and non-interchangeable after training.
For instance, $\hat y_1$ is always the prediction that corresponds to the first query in the model.
This is relevant for the application of KD\@.

For supervised training, the model is given ground truth object annotations
$
y = ((p_j, b_j))_{j\in\N}
$
where
$
p_j
$
is the indicator vector of the category of the object and $b_j \in [0,1]^4$ is its bounding box.
Images usually contain fewer objects than the number $N$ of hypotheses, so $y$ is padded with background detections for which $p_i(\phi)=1$ and $b_i$ is arbitrary.
The model is trained end-to-end to optimise the loss,
\begin{equation}\label{eq_detr}
  \mathcal{L}_\text{DETR}(\hat y, y)
  =
  \sum_{i\in\N} \langle - \log \hat p_{\hat \sigma_i}, p_i\rangle
  + \mathbf{1}_{c(p_i)\not= \phi}
  \mathcal{L}_\text{box}(\hat b_{\hat \sigma_i}, b_i),
\end{equation}
where
$
c(p_i) = \operatornamewithlimits{argmax}_{c\in\C\cup \{ \phi \}} p_i(c)
$
is the class encoded by $p_i$,
$
\mathcal{L}_\text{box}(\hat b_{\hat\sigma_i}, b_i)
=
\gamma_1 \mathcal{L}_{\text{IoU}}(\hat{b}_{\hat\sigma_i}, b_i) +
\gamma_2 \|\hat{b}_{\hat\sigma_i} - b_i\|_{1}
$
is the bounding box prediction loss and $\hat \sigma$ is the best association of ground truth labels to object hypotheses, obtained by solving the matching problem,
\begin{equation}\label{eq_matching}
\hat \sigma =
\operatornamewithlimits{argmax}_{\sigma\in\mathcal{S}_N}
\sum_{i\in\N} \mathbf{1}_{c(p_i)\not= \phi} \left \{ 
- \langle \hat p_{\sigma_i}, p_i \rangle
+
\mathcal{L}_\text{box}(\hat b_{\sigma_i}, b_i),
\right \}
\end{equation}
using the Hungarian algorithm~\cite{Kuhn1955Hungarian,Stewart2016End} (see~\cite{Carion2020DETR} for details).

\subsection{Detector knowledge distillation}%
\label{subsec_mpdistillation}

In a multi-phase learning scenario, at the beginning of a new phase, the model is initialized as $\Phi \leftarrow \Phi^\text{old}$ where $\Phi^\text{old}$ is the model trained in the phase before.
As the new data for the current phase is received, training the model $\Phi$ as normal by minimising~\cref{eq_detr} leads to forgetting.

KD~\cite{Hinton2015KD,LiH2018LwF} reduces forgetting by maintaining a copy of the old model and making sure that the outputs of the new and old models stay close.
Applied to our transformer-based detectors, given a new training image-label pair $(x,y)$, one computes the old model's output $\hat y^\text{old} = \Phi^\text{old}(x)$ and, minimizes the sum of the $\mathcal{L}_\text{DETR}(\hat y, y)$ loss with the \emph{knowledge distillation loss}
$$
\mathcal{L}_\text{KD}(\hat y, \hat y^\text{old})
=
\sum_{j\in\N}
\left[\sum_{c\in\C}
-\hat p_j(c) \log \hat p^\text{old}_j(c)
\right]
+
\mathcal{L}_\text{box}(\hat b_j, \hat b_j^\text{old}).
$$
This loss compares the output tokens of the new and old models, which makes sense since they depend on the same object queries, at least initially, and are thus in correspondence.
However, we find that this loss is dominated by background information because most of the tokens predict background.
Furthermore, transformer-based detectors aim to find one-to-one matchings between predictions and ground-truth labels without duplicates, which is not accounted for by the classical KD loss.

The key issue is that summing losses $\mathcal{L}_\text{DETR} + \mathcal{L}_\text{KD}$ as in standard KD fails to properly account for the \emph{structure} of the labels, which is crucial for detection problems, particularly in an incremental learning setting.
Specifically, the old model knows about all categories seen so far during training \emph{except} the new categories that are annotated in the current phase.
However, the new training images contain multiple objects, including the old types, which are thus \emph{not} annotated in the current phase.
This means that $\mathcal{L}_\text{DETR}$ and $\mathcal{L}_\text{KD}$ provide potentially contradictory supervision.

We thus suggest that, in a detection context, new and old knowledge should be fused in a \emph{structured manner}.
As illustrated in \cref{fig_framework_1}, we do so by selecting the most confident foreground predictions from the old model and using them as pseudo labels.
We purposefully ignore background predictions because they are imbalanced, and they can contradict the labels of the new classes available in the current phase.
Then, we \emph{merge} the pseudo labels for the old categories with the ground-truth labels for the new categories and use bipartite matching to train the model on the joint labels.
This inherits the good properties of the original formulation such as ensuring one-to-one matching between labels and hypotheses and avoiding duplicate detections.

Formally, given the predictions $\hat y^\text{old}$ from the old model, we first identify the subset $\F \subset \N$ of the ones that are predicted as foreground:
$
\F = \{
  j \in \N : \forall c \in \C: \hat p^\text{old}_j(c) > \hat p^\text{old}_j(\phi)
\}.
$
Of these, we pick the subset $\PP \subset \F$, $|\PP|=K$ formed by the $K$ most confident predictions, \ie,
$$
\forall i \in \PP,~j \in \F-\PP:~
\max_{c\in\C} \hat p^\text{old}_i(c)
>
\max_{c\in\C} \hat p^\text{old}_j(c).
$$
Finally, we further restrict the predictions to the subset $\Q \subset \PP$ that does not overlap too much with the ground-truth labels for the new categories:
$$
\Q = \{ j \in \PP :~\forall i \in \N:~
c(p_i) \not= \phi \Rightarrow \operatorname{IoU}(\hat b^\text{old}_j, b_i) \leq \lambda \}.
$$
In the experiments, we set $\lambda=0.7$.
We keep a filtered set of pseudo-labels:
\begin{equation}\label{eq_pseudo}
\hat y^\text{pseudo} = (\hat y^\text{old}_j)_{j \in \Q}.
\end{equation}

Next, we distill knowledge from the current labels $y$ and the pseudo-labels obtained from the old model into a single, coherent set of labels 
\begin{equation}\label{eq_distill}
  y^\text{distill}
  =
  (y_i)_{i:c(p_i)\not=\phi} 
  \oplus
  \hat y^\text{pseudo}
  \oplus
  y^\text{bg},
\end{equation}
where we concatenate the object labels for the new categories, the pseudo-labels, and enough background labels $y^\text{bg}$ to pad $y^\text{distill}$ to contain $N$ elements.

In this manner, the distillation occurs at the level of the labels.
The model is still trained by using \cref{eq_detr} as before, resulting in the \emph{detector knowledge distillation} (DKD) loss:
\begin{equation}\label{eq_dkd}
\mathcal{L}_\text{DKD}(\hat y, y^\text{distill})
=
\mathcal{L}_\text{DETR}(\hat y, y^\text{distill}).
\end{equation}
Besides the usage of the distilled labels, the main difference between~\cref{eq_detr,eq_dkd} is that, while the class distribution $p_i$ for the new label is deterministic, it is \emph{not} for the pseudo-labels.
Plugged in~\cref{eq_detr}, this results in the standard distillation effect for categorical distributions trained using the cross entropy loss.

\input{pseudocode/pseudocode}

\input{tables/table_sota}

\subsection{Distribution-preserving calibration}%
\label{subsec_learning_calibration}

ER methods, which store a small number of exemplars and replay them in future phases, are shown to be effective in preserving the old category knowledge in IOD~\cite{Joseph2021TowardsOpenWorld,Liu2020MultiTask}, but can suffer from the severe imbalance between old and new category annotations.
Incremental learning methods for classification~\cite{Hou2019LUCIR,Liu2020AANets,Wu2019BiC} usually use re-balancing strategies to address the imbalance problem. They create a category-balanced subset of the data and finetune some model components (\eg, the classifier) on it.
However, such strategies do not apply directly to the IOD setting.
First, the class distribution in detection is far from balanced, and a better strategy is to match the natural data distribution instead of the uniform one.
Second, because there are multiple objects in each image, it is non-trivial to create a subset of exemplar images with a set number of objects for each category.
We address these issues next.

\paragraph{Selecting exemplars to match the training distribution.}

Called during phase $i$, \cref{alg_exemplar} produces a new exemplar subset $\EE_i$ whose distribution matches as well as possible the distribution of categories in the subset $\D_i$ of the data.
This is achieved by adding to $\EE_i$ a set number $R_i$ of one exemplar $e^* \in \D_i$, one at a time, chosen by minimizing the Kullback-Leibler divergence~\cite{Kullback1951KLDivergence} between the category marginals of $\EE_i$ and $\D_i$:
\begin{equation}\label{eq_num_anno}
e^* \gets
\sum_{c\in\mathcal C_i}p_{\D_i}(c)\log p_{\EE_i\cup \{ e \} }(c),
\end{equation}
where $p_{\D}(c)$ denotes the probability of category $c$ in dataset $\D$.
Then, the overall exemplar set $\EE_{1:i} = \EE_i \cup \EE_{1:i-1}$ is obtained as the union of the new subset just found and the previous exemplar et $\EE_{1:i-1}$.
Because classes in different subsets $\D_i$ are disjoint, this also means that, by the end of the training, the distribution of classes in $\EE_{1:M}$ approximates the one of the overall training set $\D$.

\paragraph{Learning using balanced data.}

In order to use the available data as well as possible while balancing the detector $\Phi$, in each phase we update it in two steps.
In the first step, the model is trained using the DKD loss on all the available data
$
\D_i\cup\EE_{1:i-1}
$
given by the union of the current data subset $\D_i$ and the exemplar memory $\EE_{1:i-1}$ carried over the previous training phases.
In the second step, the model is fine-tuned using the new exemplar set $\mathcal{E}_{1:i}$, ignoring $\D_i$ and using only the DETR loss, using fewer data but achieving better calibration.
The overall algorithm is given in \cref{alg_cdetr}.

%% file: figures/fig_framework_1.tex
\begin{figure*}
\centering
\includegraphics[width=1.02\textwidth]{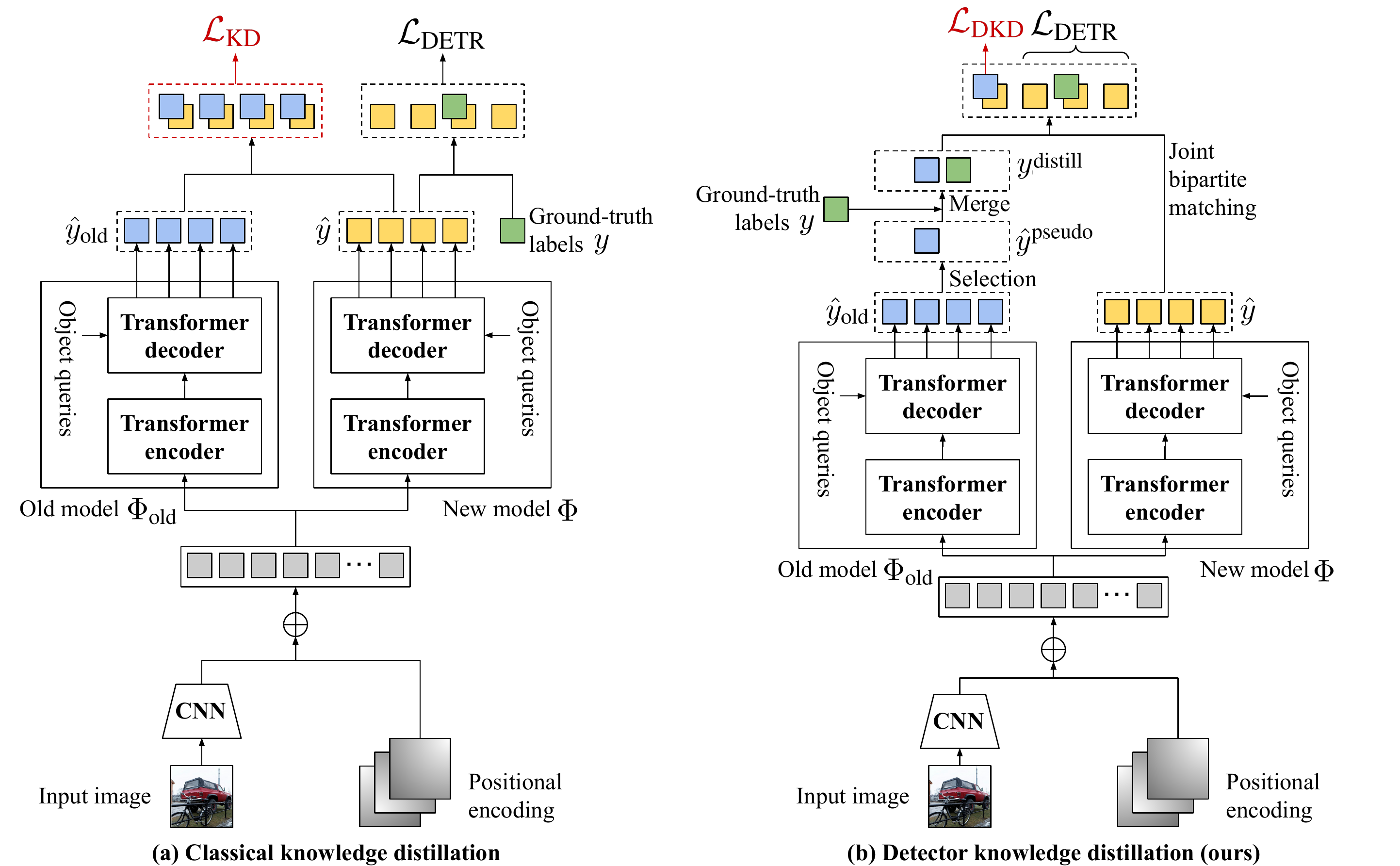}
\vspace{-0.4cm}
\caption{\textbf{(a) Classical knowledge distillation.} There are two issues when directly applying KD~\cite{Hinton2015KD,LiH2018LwF} to the transformer-based detectors~\cite{Zhu2021DeformableDETR,Carion2020DETR,Dai2021UPDETR}.
(i) Transformer-based detectors work by testing a large number of object hypotheses in parallel.
Because the number of hypotheses is much larger than the typical number of objects in an image, most of them are negative, resulting in an unbalanced KD loss.
(ii) Because both old and new object categories can co-exist in any given training image, the KD loss and regular training objective can provide contradictory evidence.
\textbf{(b) Detector knowledge distillation (ours).} We select the most confident foreground predictions from the old model and use them as pseudo labels.
We purposefully ignore background predictions because they are imbalanced and they can contradict the labels of the new classes available in the current phase. Then, we \emph{merge} the pseudo labels for the old categories with the ground-truth labels for the new categories and use bipartite matching to train the model on the joint labels. This inherits the good properties of the original formulation such as ensuring one-to-one matching between labels and hypotheses and avoiding duplicate detections.}%
\label{fig_framework_1}
\vspace{-0.3cm}
\end{figure*}

%% file: pseudocode/pseudocode.tex
\begin{algorithm}
\caption{CL-DETR (the $i$-th phase)}\label{alg_cdetr}
\SetAlgoLined
\SetKwInput{KwData}{Input}
\SetKwInput{KwResult}{Output}
    \KwData{new category data $\D_{i}$; old category exemplars $\EE_{1:i-1}$; old model $\Phi^\text{old}$.}
    \KwResult{new model $\Phi$; exemplars $\EE_{1:i}$.}
    Get $\D_{i}$ and load $\EE_{1:i-1}$ from memory\;
    Let $\Phi \gets \Phi^\text{old}$\;
    \For{\emph{epochs}}{
    \For{\emph{mini-batches} $(x, y)\in\D_{i} \cup \EE_{1:i-1}$}{
      Let $\hat{y}^{\text{old}} \gets \Phi^\text{old}(x)$\;
      Get $\hat y^\text{pseudo}$ from $\hat{y}^{\text{old}}$ and $y$ using \cref{eq_pseudo}\;
      Get $y^\text{distill}$ from $\hat y^\text{pseudo}$ and $y$ using \cref{eq_distill}\;
      Let $\hat{y} \gets \Phi(x)$\;
      Get $\hat \sigma$ by matching $y^\text{distill}$ to $\hat y$ using \cref{eq_matching}\;
      Compute $\mathcal{L}_{\text{DKD}}(\hat y, y^\text{distill})$ using \cref{eq_dkd}\;
      Update $\Phi$ via a gradient step.
      }
    }
  Build the exemplar set $\mathcal \EE_{1:i}$ using \cref{alg_exemplar}\;
  \For{\emph{epochs}}{
    \For{\emph{mini-batches} $(x, y)\in\EE_{1:i}$}{
      Let $\hat{y} \gets \Phi(x)$\;
      Compute $\mathcal{L}_{\text{DETR}}(\hat y, y)$ using \cref{eq_detr}\;
      Update $\Phi$ via a gradient step\;
      }
    }
  Save $\mathcal E_{1:i}$ to the memory.
\end{algorithm}

\begin{algorithm}
\caption{Exemplar selection (the $i$-th phase)}\label{alg_exemplar}
\SetAlgoLined
\SetKwInput{KwData}{Input}
\SetKwInput{KwResult}{Output}
\KwData{new category data $\D_{i}$; old category exemplars $\EE_{1:i-1}$; target number of exemplars ${R}_{i}$.}
\KwResult{exemplars $\EE_{1:i}$.}
Let $\EE_i\gets \{ \}$\;
\Repeat{$R_i$ times}{
    Select $e \in \D_i$ according to \cref{eq_num_anno}\;
    Let $\EE_i \gets \EE_i \cup \{x\}$\;
}
Let $\mathcal E_{1:i} \gets \mathcal \EE_{i} \cup \mathcal \EE_{1:i-1}$.
\end{algorithm}

%% file: tables/table_sota.tex
\begin{table*}
\small
\centering
\setlength{\tabcolsep}{3.25mm}{
\begin{tabular}{l|lcccccccc}
\toprule
Setting    & Method & Detection baseline & $AP$ & $AP_{50}$ & $AP_{75}$  & $AP_{S}$  & $AP_{M}$  & $AP_{L}$  \\ \midrule
     \multirow{6}{*}{$70${$+$}$10$} & ERD~\cite{Feng2022ElasticResponse} & UP-DETR    & 36.2\tiny{$\pm$0.3} & 54.8\tiny{$\pm$0.4}  & 39.3\tiny{$\pm$0.4} & 20.8\tiny{$\pm$0.3}  & 39.3\tiny{$\pm$0.5} & 47.9\tiny{$\pm$0.3}  \\
     & \cellcolor{mygray-bg}{CL-DETR (ours)} & \cellcolor{mygray-bg}{UP-DETR}  &  \cellcolor{mygray-bg}{{37.6}\tiny{$\pm$0.2}} & \cellcolor{mygray-bg}{{56.5}\tiny{$\pm$0.4}}  & \cellcolor{mygray-bg}{{39.4}\tiny{$\pm$0.3}} & \cellcolor{mygray-bg}{{20.5}\tiny{$\pm$0.3}}  & \cellcolor{mygray-bg}{{39.1}\tiny{$\pm$0.4}} & \cellcolor{mygray-bg}{{49.9}\tiny{$\pm$0.3}} \\
     & LwF~\cite{LiH2018LwF} & Deformable DETR   & 24.5\tiny{$\pm$0.3} & 36.6\tiny{$\pm$0.2}  & 26.7\tiny{$\pm$0.4} & 12.4\tiny{$\pm$0.2}  & 28.2\tiny{$\pm$0.4} & 35.2\tiny{$\pm$0.4} \\
     & iCaRL~\cite{Rebuffi2017iCaRL} & Deformable DETR   & 35.9\tiny{$\pm$0.4} & 52.5\tiny{$\pm$0.3}  & 39.2\tiny{$\pm$0.3} & 19.1 \tiny{$\pm$0.3} & 39.4\tiny{$\pm$0.5} & 48.6\tiny{$\pm$0.3} \\
     & ERD~\cite{Feng2022ElasticResponse} & Deformable DETR    & 36.9\tiny{$\pm$0.4} & 55.7\tiny{$\pm$0.4}  & 40.1\tiny{$\pm$0.4} & 21.4\tiny{$\pm$0.3}  & 39.6\tiny{$\pm$0.3} & 48.7\tiny{$\pm$0.3}  \\
     & \cellcolor{mygray-bg}{CL-DETR (ours)} & \cellcolor{mygray-bg}{Deformable DETR} &  \cellcolor{mygray-bg}{\textbf{40.1}\tiny{$\pm$0.3}} & \cellcolor{mygray-bg}{\textbf{57.8}\tiny{$\pm$0.4}}  & \cellcolor{mygray-bg}{\textbf{43.7}\tiny{$\pm$0.3}} & \cellcolor{mygray-bg}{\textbf{23.2}\tiny{$\pm$0.3}}  & \cellcolor{mygray-bg}{\textbf{43.2}\tiny{$\pm$0.2}} & \cellcolor{mygray-bg}{\textbf{52.1}\tiny{$\pm$0.3}} \\
     \\[-12pt]
     \midrule
     \multirow{6}{*}{$40${$+$}$40$} & ERD~\cite{Feng2022ElasticResponse} & UP-DETR   & 35.4\tiny{$\pm$0.4} & 55.1\tiny{$\pm$0.3}  & 38.3\tiny{$\pm$0.3} & 17.9\tiny{$\pm$0.4}  & 39.0\tiny{$\pm$0.3} & 49.8\tiny{$\pm$0.3} \\
     & \cellcolor{mygray-bg}{CL-DETR (ours)} & \cellcolor{mygray-bg}{UP-DETR}  &  \cellcolor{mygray-bg}{{37.0}\tiny{$\pm$0.2}} & \cellcolor{mygray-bg}{\textbf{56.2}\tiny{$\pm$0.2}}  & \cellcolor{mygray-bg}{{39.1}\tiny{$\pm$0.4}} & \cellcolor{mygray-bg}{\textbf{20.9}\tiny{$\pm$0.2}}  & \cellcolor{mygray-bg}{{38.9}\tiny{$\pm$0.3}} & \cellcolor{mygray-bg}{{49.2}\tiny{$\pm$0.3}} \\
     & LwF~\cite{LiH2018LwF} & Deformable DETR   & 23.9\tiny{$\pm$0.2} & 41.5\tiny{$\pm$0.3} & 25.0\tiny{$\pm$0.3} & 12.0\tiny{$\pm$0.4}  & 26.4\tiny{$\pm$0.3} & 33.0\tiny{$\pm$0.5} \\
     & iCaRL~\cite{Rebuffi2017iCaRL} & Deformable DETR   & 33.4\tiny{$\pm$0.4} & 52.0\tiny{$\pm$0.3}  & 36.0\tiny{$\pm$0.2} & 18.0\tiny{$\pm$0.3} & 36.4\tiny{$\pm$0.3} & 45.5\tiny{$\pm$0.4} \\
     & ERD~\cite{Feng2022ElasticResponse} & Deformable DETR   & 36.0\tiny{$\pm$0.2} & 55.2\tiny{$\pm$0.2}  & 38.7\tiny{$\pm$0.3} & 19.5\tiny{$\pm$0.2}  & 38.7\tiny{$\pm$0.3} & 49.0\tiny{$\pm$0.4} \\
     & \cellcolor{mygray-bg}{CL-DETR (ours)} & \cellcolor{mygray-bg}{Deformable DETR} &  \cellcolor{mygray-bg}{\textbf{37.5}\tiny{$\pm$0.3}} & \cellcolor{mygray-bg}{{55.1}\tiny{$\pm$0.4}}  & \cellcolor{mygray-bg}{\textbf{40.3}\tiny{$\pm$0.2}} & \cellcolor{mygray-bg}{\textbf{20.9}\tiny{$\pm$0.2}}  & \cellcolor{mygray-bg}{\textbf{40.8}\tiny{$\pm$0.4}} & \cellcolor{mygray-bg}{\textbf{50.7}\tiny{$\pm$0.2}} \\
     [-2pt]
\bottomrule 
\end{tabular}
}
\vspace{-0.1cm}
\caption{IOD results (\%) on COCO 2017.
In the $A+B$ setup, in the first phase, we observe a fraction $\frac{A}{A+B}$ of the training samples with $A$ categories annotated.
Then, in the second phase, we observe the remaining $\frac{B}{A+B}$ of the training samples, where $B$ new categories are annotated.
We test settings $A+B=40+40$ and $70+10$.
Exemplar replay is applied for all methods except for LwF~\cite{LiH2018LwF}.
We run experiments for three different categories and data orders and report the average AP with $95\%$ confidence interval.
}
\label{table_sota}
\vspace{-0.1em}
\end{table*}

%% file: sections/4_experiments.tex
\section{Experiments}%
\label{sec_exp}

\input{figures/fig_multiple_phase}

\input{tables/table_ablation}

We evaluate CL-DETR on COCO 2017 using two transformer-based detectors, Deformable DETR and UP-DETR~\cite{Zhu2021DeformableDETR,Dai2021UPDETR} as the baselines and achieve consistent improvements compared to the baselines and a direct application of KD and ER\@.
Below we describe the dataset and implementation details (\cref{subsec_datasets}) followed by results and analyses (\cref{subsec_results}).

\subsection{Dataset and implementation details}%
\label{subsec_datasets}

\paragraph{Dataset and evaluation metrics.}

We conduct IOD experiments on COCO 2017~\cite{Lin2014COCO}, which is widely used in related works~\cite{Feng2022ElasticResponse,Peng2021SID,Zhu2021DeformableDETR,Dai2021UPDETR}.
Following~\cite{Feng2022ElasticResponse}, the standard COCO metrics are used for evaluation, \ie, $AP$, $AP_{50}$, $AP_{75}$, $AP_S$, $AP_M$, and $AP_L$.
In the ablation study, we introduce a new metric, \emph{forgetting percentage points} (FPP), measuring the difference between the AP of the first and last phase models on the categories observed in the first phase.

\paragraph{Experiment setup.}

We conduct IOD experiments in the following setting.
\textbf{\emph{Two-phase setting:}}
In the $A+B$ setup, in the first phase, we observe a fraction $\frac{A}{A+B}$ of the training samples with $A$ categories annotated.
Then, in the second phase, we observe the remaining $\frac{B}{A+B}$ of the training samples, where $B$ new categories are annotated.
We test settings $A+B=40+40$ and $70+10$.
\textbf{\emph{Multiple-phase setting:}}
In the $40+X\times Y$ setup, in the first phase, we observe half of the training samples with $40$ categories annotated.
In each following phase, we observe $\frac{1}{2Y}$ of the training samples we have never seen before with annotations for $X$ new categories.
We run experiments for $40+20\times2$ and $40+10\times4$.
We repeat each experiment three times, randomizing the order of categories and data in the different phases, and report the average APs.
The total memory budget for the exemplars is set as $10\%$ of the total dataset size.

\paragraph{Implementation details.}

We follow~\cite{Zhu2021DeformableDETR,Dai2021UPDETR} and use an ImageNet pre-trained ResNet-50 backbone.
For the experiments on Deformable DETR~\cite{Zhu2021DeformableDETR}, we use the standard configurations without their iterative bounding box refinement mechanism and the two-stage Deformable DETR\@.
We train the model for $50$ (Deformable DETR) and $150$ epochs (UP-DETR), following the original implementations~\cite{Zhu2021DeformableDETR,Dai2021UPDETR}.
In order to apply our distribution-preserving calibration (\cref{subsec_learning_calibration}), we train the coarse Deformable DETR (UP-DETR) model for $40$ ($120$) epochs and perform calibration for $10$ ($30$) epochs to preserve the total number of epochs. 

\subsection{Results and analyses}%
\label{subsec_results}

\paragraph{Two-phase setting.}

\Cref{table_sota} shows that, in the two-phase settings $70+10$ and $40+40$, applying CL-DETR to Deformable DETR~\cite{Zhu2021DeformableDETR} and UP-DETR~\cite{Dai2021UPDETR} consistently performs better than the state-of-the-art~\cite{Feng2022ElasticResponse} and other IOD methods~\cite{LiH2018LwF,Rebuffi2017iCaRL}.
In particular, Deformable DETR~\cite{Zhu2021DeformableDETR} \emph{w/} ours achieves the highest AP, \eg, $40.1\%$ and $37.5\%$ in the $70+10$ and $40+40$ settings, respectively.
The performance gap is larger with more categories in the $1$-st phase.
\Eg, the AP differences between our method and~\cite{Feng2022ElasticResponse} are $3.2$ and $1.5$ percentage points when we observe $70$ and $40$ categories in the first phase, respectively, likely due to CL-DETR benefiting more from a well-pre-trained model.  

\paragraph{Multiple-phase setting.}

\Cref{figure_acc_plots} evaluates CL-DETR in the multiple-phase setting
with large gains compared to other IOD methods in both the $40+20\times2$ and $40+10\times4$ experimental variants.
The relative advantage of CL-DETR increases with the number of phases.
For instance, our method improves the AP of~\cite{Feng2022ElasticResponse} by $2.9$ percentage points in the $40+20\times2$ setting and by $7.4$ percentage points in the $40+10\times4$ setting.
This suggests that the advantage of CL-DETR shows more in challenging settings, where the forgetting problem is stronger due to the larger number of training phases.

\paragraph{Ablation study for DKD.}

In \cref{table_ablation} (Rows 1--4) we ablate our DKD approach\@.
By comparing row~2 to row~1, we observe that classical KD significantly improves the IOD performance compared to the baseline (\ie, finetuning the model without IOD techniques), but still results in large overall forgetting: $19.3$ FPP\@.
Comparing row~3 to row~2, we can see that joint bipartite matching works well and boosts the AP of all categories by $5.8$ percentage points compared to conventional KD\@.
The reason is that joint bipartite matching helps ensure a one-to-one matching between objects and hypotheses and discourages duplicate detections.
Comparing row~4 to row~3, our pseudo label selection further improves the AP and reduces forgetting, helping the model to ignore the redundant background information and reducing conflicts between old and new labels.

\input{tables/table_topk}

\paragraph{Ablation study for ER.}

In \cref{table_ablation} (Rows 5--6), we ablate our ER method.
Comparing row 6 to row 5, we can see that the calibration strategy of~\cref{subsec_learning_calibration} boosts both the all-category and old-category performance, by $1.8$ and $2.1$ percentage points respectively, compared to using conventional ER~\cite{Rebuffi2017iCaRL,Liu2020MultiTask}.
It also helps to overcome the catastrophic forgetting problem in IOD, reducing the AP forgetting by $2.1$ percentage points.
This is because the conventional ER balances the sample distributions, changing the category distribution of the training set, whereas our method preserves it, thus improving performance.

\paragraph{Ablation study for pseudo label selection strategies.}

In \cref{table_ablation_topk}, we show the results for two pseudo-label selection strategies:
(1) selecting top-$K$ most-confident non-background predictions (Rows 1--3); and
(2) selecting the predictions using a threshold for the prediction scores (Rows 4--6).
We observe the first strategy works better, with peak AP when $K$=$10$. The maximum performance difference is only $0.4$ percentage points when using different values for $K$\@.
This indicates our method is robust to its hyperparameter settings.

\input{figures/fig_visual}

\paragraph{Visualizations.}

\Cref{fig_visual} visualizes the old category pseudo (blue) and ground-truth (green) bounding boxes in some training samples in COCO 2017.
In \cref{fig_visual}~(a,~b), CL-DETR generates accurate pseudo bounding boxes that exactly match the ground-truth ones.
This shows the effectiveness of our pseudo-label selection strategy.
In \cref{fig_visual}~(c,~d), CL-DETR fails to generate pseudo bounding boxes for all objects in the images when there are too many.
This is explained by our strategy of selecting the top-$K$ most-confident non-background bounding boxes as the pseudo-labels followed by removing the ones that overlap with the new category ground-truth labels excessively.
In this manner, the number of pseudo bounding boxes is always smaller than $K$.
The trade-off, justified by our improvements in the experiments, is to prefer correct although possibly incomplete annotations to contradictory or noisy ones.

%% file: figures/fig_multiple_phase.tex
\begin{figure*}
\newcommand{\newincludegraphics}[1]{\includegraphics[height=1.45in]{#1}}
\centering
\includegraphics[height=0.20in]{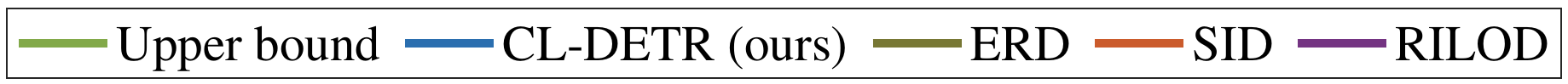}\\
\vspace{0.2cm}
\newincludegraphics{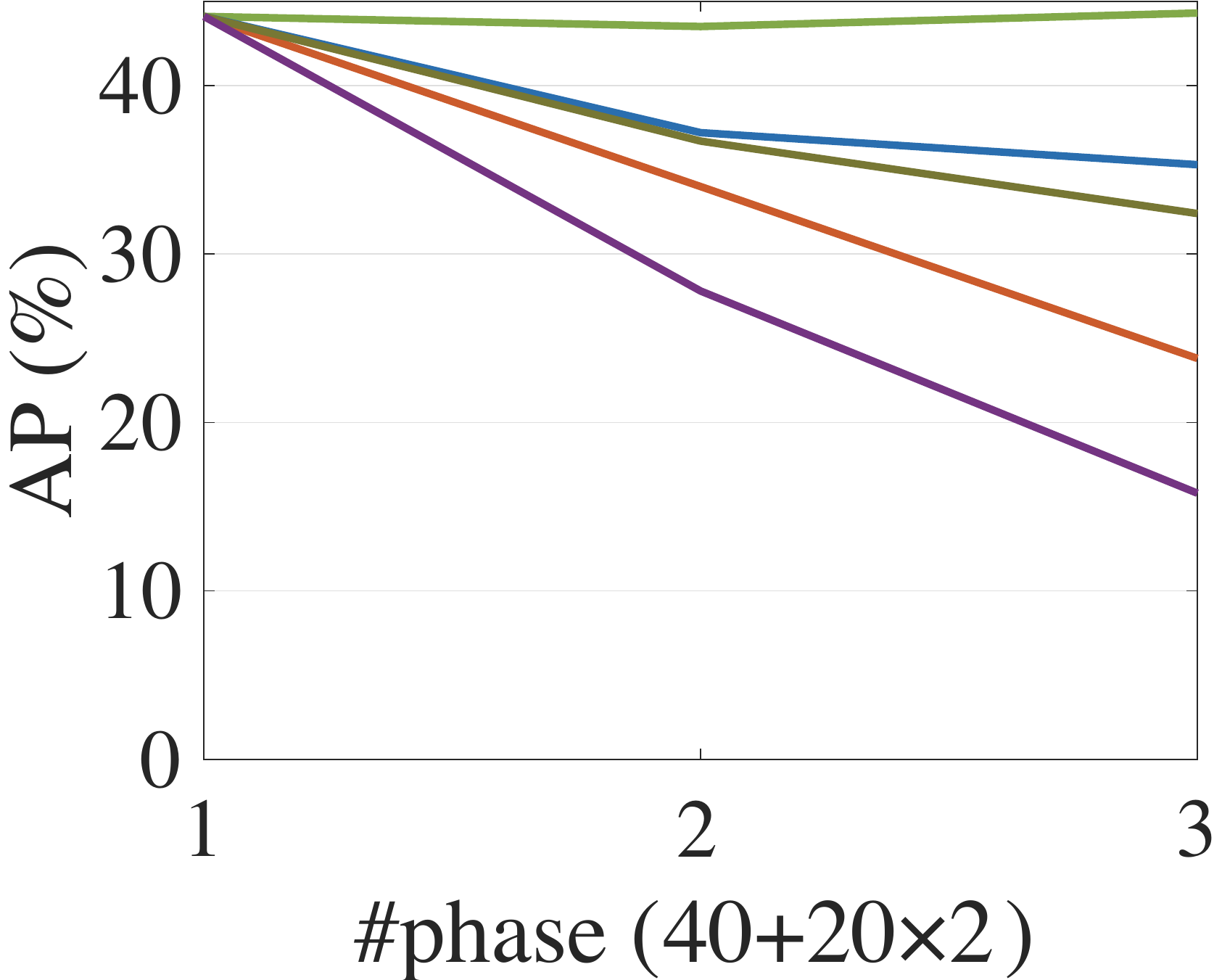}
\newincludegraphics{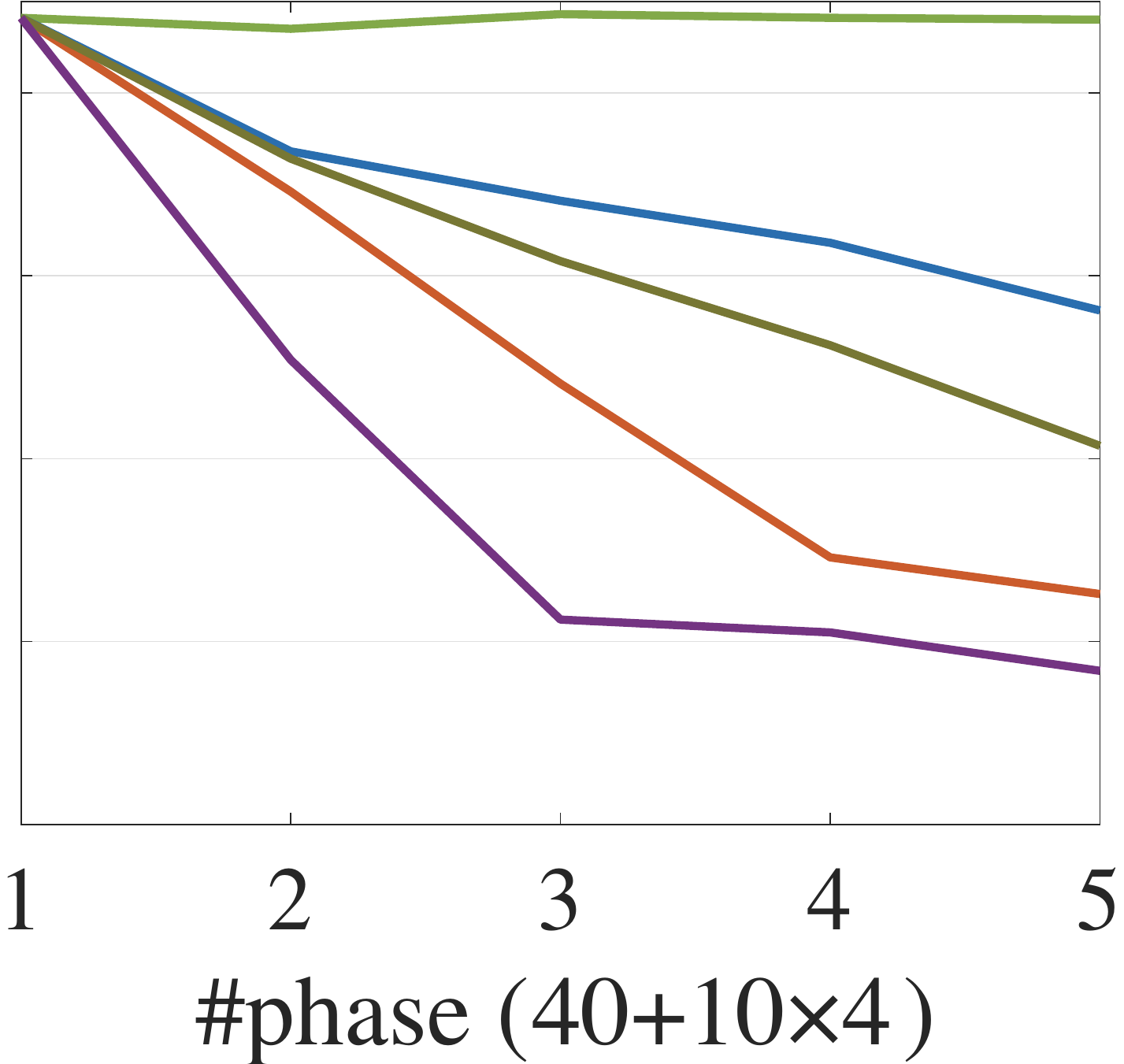}
\hspace{1mm}
\newincludegraphics{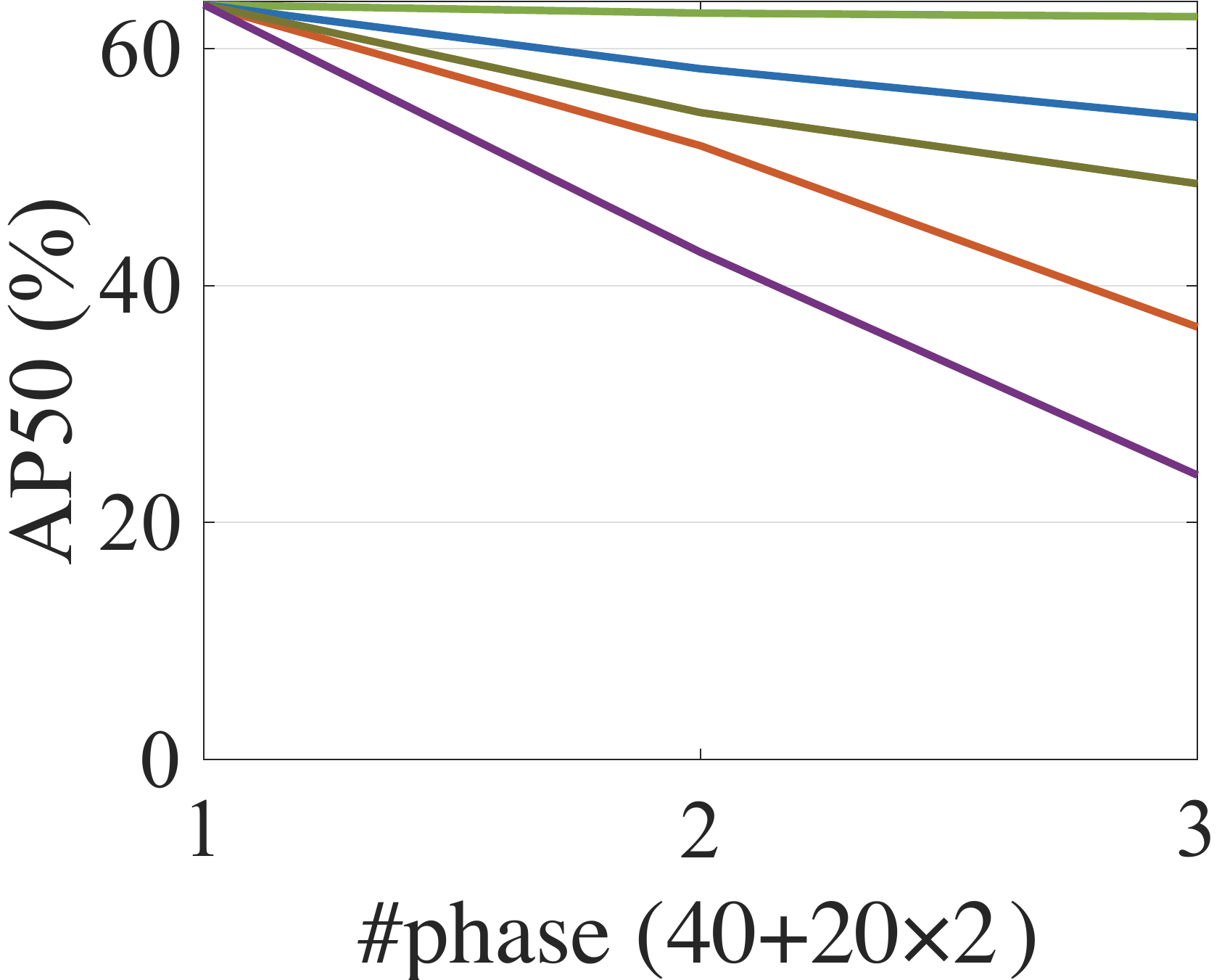}
\newincludegraphics{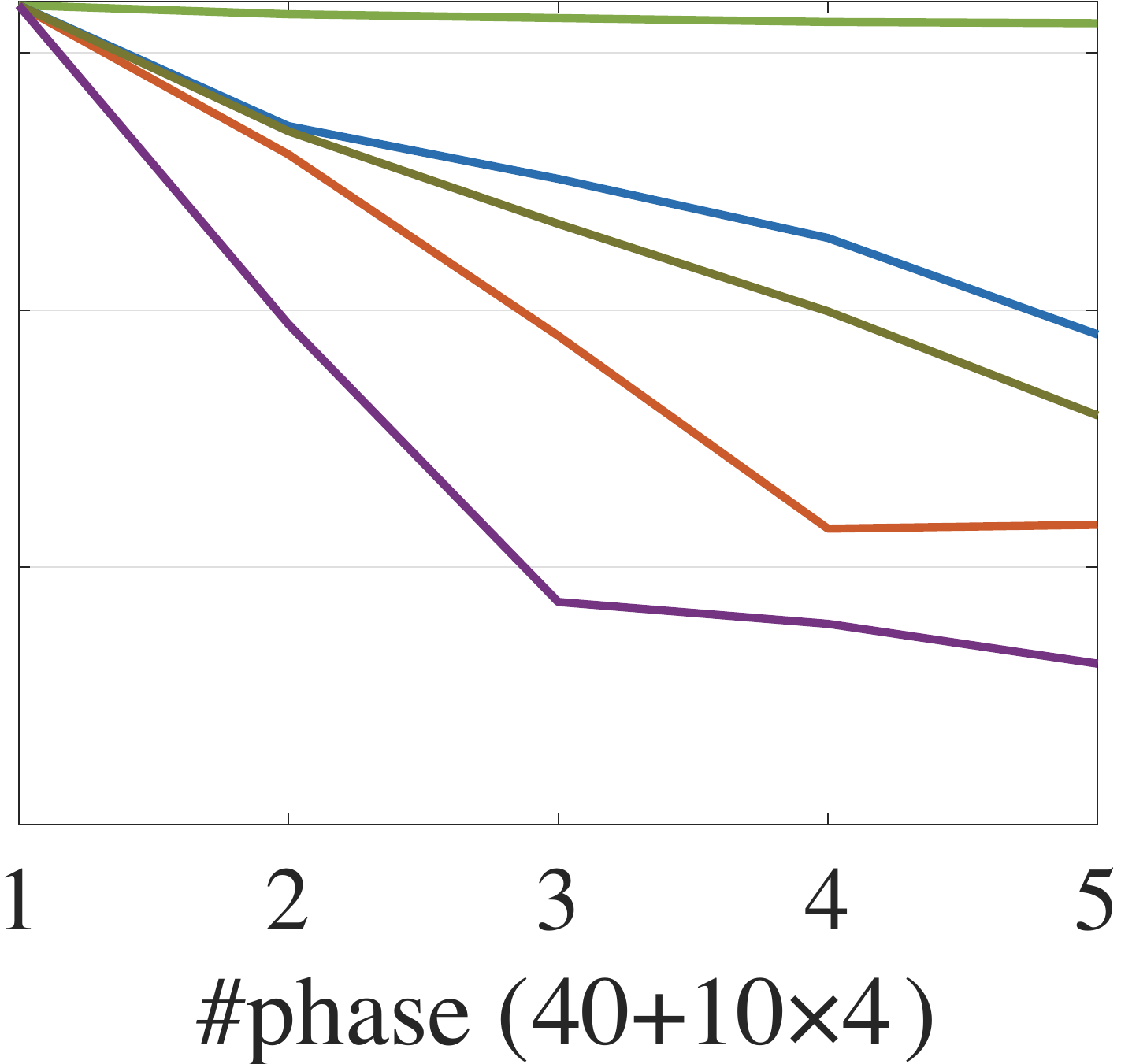}
\vspace{-0.1cm}
\caption{IOD results (AP/AP$_{50}$, \%) on COCO 2017 in the $40+20\times2$ and $40+10\times4$ settings. 
Our method is based on Deformable DETR\@.
Comparing methods: Upper Bound (the results of joint training with all previous data accessible in each phase), ERD~\cite{Feng2022ElasticResponse}, SID~\cite{Peng2021SID}, and RILOD~\cite{Li2019RILOD}.
The results of the related works are from~\cite{Feng2022ElasticResponse}.
We use the same data split as~\cite{Feng2022ElasticResponse} for a fair comparison.
}
\label{figure_acc_plots}
\vspace{0.2cm}
\end{figure*}

%% file: tables/table_ablation.tex
\begin{table*}
\small
\centering
\setlength{\tabcolsep}{0.70mm}{
\begin{tabular}{@{}c|ccc|cc|cccccccccccccc@{}}
\toprule
  \multirow{3}{*}{{Row}} & Knowledge & Joint & Pseudo & Exemplar& Distribution& \multicolumn{4}{c}{\multirow{1.5}{*}{All categories $\uparrow$}}  && \multicolumn{4}{c}{\multirow{1.5}{*}{Old categories $\uparrow$}} && \multicolumn{4}{c}{\multirow{1.5}{*}{FPP $\downarrow$}}\\
  & distillation & bipartite & label & replay & preserving & \multirow{2.5}{*}{$AP$} & \multirow{2.5}{*}{$AP_{S}$}  & \multirow{2.5}{*}{$AP_{M}$}  & \multirow{2.5}{*}{$AP_{L}$} && \multirow{2.5}{*}{$AP$} & \multirow{2.5}{*}{$AP_{S}$}  & \multirow{2.5}{*}{$AP_{M}$}  & \multirow{2.5}{*}{$AP_{L}$} && \multirow{2.5}{*}{$AP$} & \multirow{2.5}{*}{$AP_{S}$}  & \multirow{2.5}{*}{$AP_{M}$}  & \multirow{2.5}{*}{$AP_{L}$} \\
  & (KD) & matching & selection & (ER) & calibration \\
    \midrule
    {1}& & & & & &  4.2 & 1.6  & 4.7 &5.8 && 0.7 & 0.2 & 0.8 & 0.8 && 42.6 & 25.6 & 45.1 & 56.7    \\
    {2}& \checkmark & & & & & 24.5 & 12.4  & 28.2 & 35.2 && 24.0 & 12.3 & 27.7 &34.4 && 19.3 & 13.5 & 18.2 & 23.1\\
    {3} & \checkmark & \checkmark & & & & 30.3 & 19.5  & 33.0 & 39.0 && 33.4 & 21.8 & 36.4 & 43.2 && 9.9 &4.0 & 9.5 & 14.3 \\
    {4}& \checkmark & \checkmark & \checkmark & &  & 33.9 & 16.3  & 37.1 & 49.2 && 33.9  & 16.6 & 36.8 & 50.0 && 9.4 & 9.2 & 9.1 & 7.5\\
    \midrule
    {5} & \checkmark & \checkmark & \checkmark & \checkmark & & 37.9 & 20.8  & 40.9 & 50.4 && 39.0 & 21.6 & 41.7  & 52.3 && 4.3 & 4.2 & 4.2 & 5.2 \\
    \cellcolor{mygray-bg}{{6}} & \cellcolor{mygray-bg}{\checkmark} & \cellcolor{mygray-bg}{\checkmark} & \cellcolor{mygray-bg}{\checkmark} &\cellcolor{mygray-bg}{} &\cellcolor{mygray-bg}{\checkmark} & \cellcolor{mygray-bg}{\textbf{40.1}} & \cellcolor{mygray-bg}{\textbf{23.2}}  & \cellcolor{mygray-bg}{\textbf{43.2}} & \cellcolor{mygray-bg}{\textbf{52.1}} &\cellcolor{mygray-bg}{}& \cellcolor{mygray-bg}{\textbf{41.8}} &\cellcolor{mygray-bg}{\textbf{24.5}} &\cellcolor{mygray-bg}{\textbf{44.7}} & \cellcolor{mygray-bg}{\textbf{54.6}} &\cellcolor{mygray-bg}{}& \cellcolor{mygray-bg}{\textbf{1.5}} & \cellcolor{mygray-bg}{\textbf{1.3}} & \cellcolor{mygray-bg}{\textbf{1.2}} & \cellcolor{mygray-bg}{\textbf{2.9}}\\
    [-2pt]
\bottomrule
\end{tabular}
}
\vspace{-0.1cm}
\caption{Ablation results (\%) for KD and ER, using Deformable DETR~\cite{Zhu2021DeformableDETR} on COCO 2017 in the $70+10$ setting.
``All categories'' (higher is better) denote the results of the last phase model on $80$ categories.
``Old categories'' (higher is better) denote the results of the last phase model on $70$ categories observed in the first phase.
``Forgetting percentage points (FPP)'' (lower is better) show the difference between the AP of the first-phase model and the last-phase model on $70$ categories observed in the first phase.
The baseline (row~1) is finetuning the model without IOD techniques. Our method (CL-DETR) is shown in row~6.}
\label{table_ablation}
\end{table*}

%% file: tables/table_topk.tex
\begin{table}
\normalsize
\begin{center}
\renewcommand\arraystretch{1.2}
\setlength{\tabcolsep}{1.3mm}{
\begin{tabular}{clcccccc}
\toprule
Row & Setting & $AP$ & $AP_{50}$  & $AP_{75}$ & $AP_{S}$  & $AP_{M}$  & $AP_{L}$  \\
\midrule
{1} &$K$=$5$& 39.7 & 57.4 & 43.1 & 22.7 & 42.6 & \textbf{52.7}\\
\cellcolor{mygray-bg}{{2}}& \cellcolor{mygray-bg}{$K$=$10$}& \cellcolor{mygray-bg}{\textbf{40.1}} & \cellcolor{mygray-bg}{\textbf{57.8}} & \cellcolor{mygray-bg}{\textbf{43.7}} & \cellcolor{mygray-bg}{23.2} & \cellcolor{mygray-bg}{\textbf{43.2}} & \cellcolor{mygray-bg}{52.1}\\
{3} &$K$=$20$& 39.9 & 57.8 & 43.2 & \textbf{23.5} & 42.9 & 51.7\\
\midrule
{4} &$p${$\geq$}$0.1$& 39.3 & 57.1 & 42.9 & 22.6 &42.3 & 52.5\\
{5} &$p${$\geq$}$0.3$& 39.6 & 57.5 & 43.0 & 23.2 &42.4 & 52.2\\
{6} &$p${$\geq$}$0.5$& 39.2 & 56.8 & 42.4 & 22.3 & 41.9 & 51.8\\
\bottomrule
\end{tabular}}
\end{center}
\vspace{-0.5cm}
\caption{Ablation result ($\%$) for different pseudo label selection strategies on COCO 2017 using the $70+10$ setting.
Rows 1--3 show the results for using different $K$ when selecting top-$K$ most-confident non-background predictions.
Rows 4--6 show the results for using different thresholds $p$ of the prediction scores to select the non-background predictions.}
\label{table_ablation_topk}
\vspace{-0.2cm}
\end{table}

%% file: figures/fig_visual.tex
\begin{figure}
\centering
\includegraphics[width=0.48\textwidth]{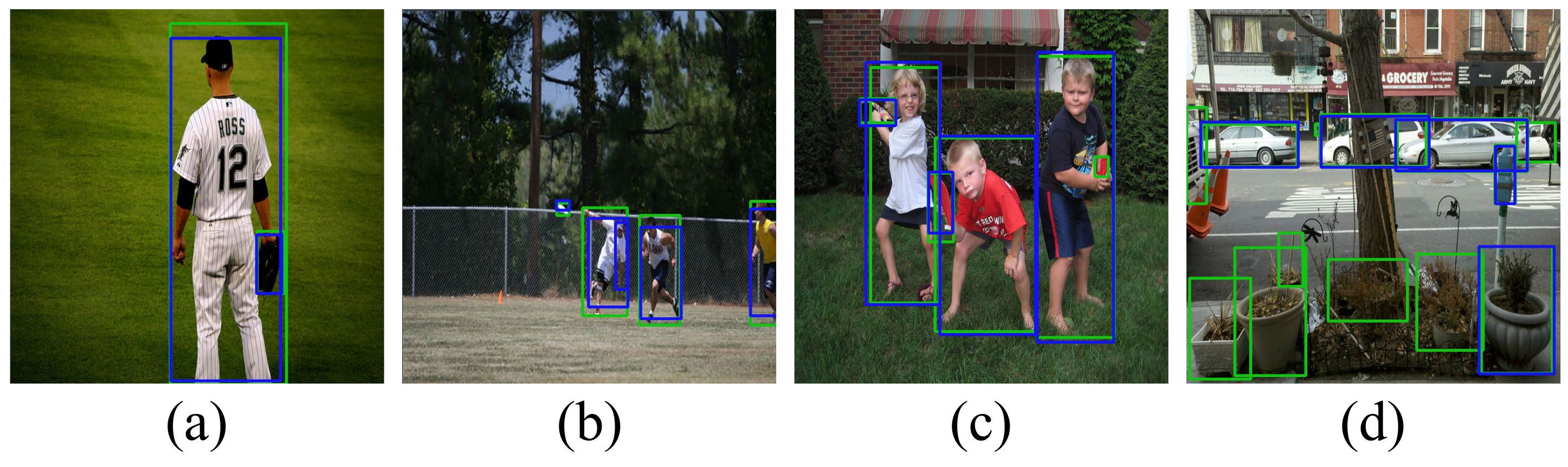}
\vspace{-0.6cm}
\caption{Visualizations of the old category pseudo (\textcolor[rgb]{0.08, 0.08, 0.95}{blue}) and ground-truth (\textcolor[rgb]{0.08, 0.77, 0.08}{green}) bounding boxes on COCO 2017 using the $70+10$ setting. \textbf{(a,~b)}:~Our method generates accurate pseudo bounding boxes that exactly match the ground-truth ones. \textbf{(c,~d)}:~When there are too many annotations in the images, generated pseudo bounding boxes cannot cover all ground-truth ones. However, the pseudo bounding boxes are still focused on the foreground objects.
}
\label{fig_visual}
\vspace{-0.2cm}
\end{figure}

%% file: sections/5_conclusion.tex
\section{Conclusions}%
\label{sec_conclusions}

This paper introduced CL-DETR, a novel IOD method that can effectively use KD and ER in transformer-based detectors.
CL-DETR improves the standard KD loss by introducing DKD which selects the most informative predictions from the old model, rejecting redundant background predictions, and ensuring that the distilled information is consistent with the new ground-truth evidence.
CL-DETR also improves ER by selecting exemplars to match the distribution of the training set. 
CL-DETR is fairly generic and can be easily applied to different transformer-based detectors, including Deformable DETR~\cite{Zhu2021DeformableDETR} and UP-DETR~\cite{Dai2021UPDETR}, achieving large improvements.
We have also defined a more realistic IOD benchmark protocol that avoids using duplicated images in different training phases.
In the future, we plan to extend our method to more challenging settings such as online learning.

\paragraph{Ethics.}

We use the COCO dataset in a manner compatible with their terms; this data contains personal information (faces).
For further details on ethics, data protection, and copyright please see \url{https://www.robots.ox.ac.uk/~vedaldi/research/union/ethics.html}.

\paragraph{Acknowledgments.}

C\@. R\@. and A\@. V\@. are supported by ERC-UNION-CoG-101001212.
C\@. R\@. is also supported by VisualAI EP/T028572/1.

%% file: supplementary/supplementary.tex
\beginsupp
\setcounter{section}{0}
\renewcommand\thesection{\Alph{section}}
\noindent

\noindent
{\Large {\textbf{Supplementary materials}}}

\vspace{0.5cm}

We present the following supplementary content: results with the traditional IOD benchmark protocol (\S\ref{supp_sec_traditional_protocol}), more ablation results (\S\ref{suppsec_ablation_study}), more visualization results (\S\ref{suppsec_visualiztion}), and instructions for our code (\S\ref{suppsec_code}).

\section{Traditional IOD protocol and results}
\label{supp_sec_traditional_protocol}

\myparagraphsupp{This is supplementary to Section~{4.2}.}
In previous work~\cite{Feng2022ElasticResponse}, in each phase, the incremental detector is allowed to observe all images that contain a certain type of objects.
Because images often contain a mix of object classes, both old and new, this means that the same images can be observed in different training phases.
This is incompatible with the standard definition of incremental learning~\cite{Rebuffi2017iCaRL,Liu2020Mnemonics,Hou2019LUCIR} where, with the exception of the examples deliberately stored in the exemplar memory, the images observed in different phases do not repeat. Thus, we provide results with our new IOD benchmark protocol in the main paper. 

For completeness and comparison, here we also evaluate performance using the traditional IOD benchmark protocol used in~\cite{Feng2022ElasticResponse}, and provide comparison results between our method and other top-performing IOD methods with this protocol.   

\paragraph{Traditional IOD protocol.}
Formally, let $\D = \{(x,y)\}$ be a dataset of images $x$ with corresponding object annotations $y$, such as COCO~2017~\cite{Lin2014COCO}, and let $\C=\{1,\dots,C\}$ be the set of object categories.
We adopt such a dataset for benchmarking IOD as follows.
First, we partition $\C$ into $M$ subsets $\C=\C_1\cup\dots\cup\C_M$, one for each training phase.
For each phase $i$, we modify the samples $(x,y)\in\D$, where $y$ only contains annotations for objects of class $\C_i$ and drop the others.
In phase $i$ of training, the model is only allowed to observe images that contain at least one annotation for objects of types $\C_i \subset\C$.

\paragraph{Experiment results.} \Cref{table_sota_supp} shows that also with the traditional IOD protocol our CL-DETR consistently performs better than the state-of-the-art~\cite{Feng2022ElasticResponse} and other IOD methods~\cite{LiH2018LwF,Li2019RILOD,Peng2021SID}. Interestingly, our method achieves better performance than other methods~\cite{Feng2022ElasticResponse,LiH2018LwF,Li2019RILOD,Peng2021SID} even without using exemplars. For example, the AP of our CL-DETR \emph{w/o} ER is $2.3$ percentage points higher than the AP of ERD~\cite{Feng2022ElasticResponse} in the $40+40$ setting. 

\input{supplementary/tables/supp_table_sota.tex}

\section{More ablation results}
\label{suppsec_ablation_study}

\myparagraphsupp{This is supplementary to Section~{4.2}.} 

\paragraph{Ablation results for $\lambda$.} 
In \cref{table_rebuttal_1}, we show the ablation results for $\lambda$ on COCO 2017 in the 70+10 setting. We can observe the peak AP is at $\lambda$=$0.7$, with a maximum performance difference of only $1.0$ percentage points using different values. This demonstrates the robustness of our method to different $\lambda$ values. Further results and analysis will be included in the final paper.

\input{supplementary/tables/supp_table_lambda}

\paragraph{Separate validation sets.} In \cref{table_rebuttal_4}, we provide ablation results for different pseudo label selection strategies on a separate validation set (COCO 2017, $70$+$10$ setting). Results show that the ``top-K selection'' strategy performs best, consistent with the findings in the main paper.

\input{supplementary/tables/supp_table_separate}

\paragraph{Iteratively improves detection.} We apply curriculum learning~\cite{Bengio2009Curriculum} for the hyperparameter, $p$, i.e., decreasing $p$ from $0.5$ to $0.1$ during the training. This way, the loss for objects with low confidence will be ignored in the beginning and only included later when the model becomes more stable. \Cref{table_rebuttal_5} shows the results on COCO 2017 in the $70$+$10$ setting. Curriculum learning for $p$ slightly improves ($+0.3$ AP) the final performance.

\input{supplementary/tables/supp_table_curriculum}

\paragraph{More fine-grained ablation results.} \Cref{table_rebuttal_6} presents partial fine-grained results using Deformable DETR on COCO 2017 in the $70$+$10$ setting. Results show that our method, CL-DETR, outperforms related methods such as LwF and iCaRL in terms of AP, old category AP, and FPP. These results highlight the effectiveness of our CL-DETR in addressing the forgetting problem.

\input{supplementary/tables/supp_table_fine_grained}

\paragraph{Different exemplar replay methods.} In \cref{table_supp_ablation}, we provide ablation results for different exemplar replay methods. Our ``distribution-persevering'' exemplar replay strategy achieves better performance (higher AP and lower FPP) compared to the existing strategies in the related works~\cite{Rebuffi2017iCaRL,Yang2022Multi}. This shows that creating an exemplar set that follows the natural data distribution of COCO 2017 improves the results, compared to existing strategies that try to select a category-balanced subset of the data as the exemplar set and thus change the original data distribution. 

\input{supplementary/tables/supp_table_ablation.tex}

\section{More visualization results}
\label{suppsec_visualiztion}

\myparagraphsupp{This is supplementary to Section~{4.2}.}
\Cref{supp_fig_pairs} visualizes the one-to-one matching between the merged bounding boxes and new model predictions (yellow) in some training samples in COCO 2017. The merged bounding boxes include the old category pseudo (blue) and new category ground-truth (green) bounding boxes. We can observe that the old category pseudo and new category ground-truth bounding boxes are complementary and indicate the old and new category objects, respectively. It shows our method successfully resolves conflicts between pseudo and ground-truth bounding boxes and ensures the model ignores background predictions.

\section{Source Code in PyTorch}
\label{suppsec_code}

In the following, we introduce how to install the environment and run the code.

\paragraph{Installation.}
To run this project, please install Python 3.7 with Anaconda.
\begin{lstlisting}[frame=shadowbox]
conda create -n cl_detr python=3.7
\end{lstlisting}

\noindent Activate the environment as follows,
\begin{lstlisting}[frame=shadowbox]
conda activate cl_detr
\end{lstlisting}

\noindent Install PyTorch and torchvision. For example, for CUDA version is 9.2, install PyTorch and torchvision as follows,
\begin{lstlisting}[frame=shadowbox]
conda install pytorch=1.5.1 torchvision=0.6.1 cudatoolkit=9.2 -c pytorch
\end{lstlisting}

\noindent We install other requirements as follows,
\begin{lstlisting}[frame=shadowbox]
pip install -r requirements.txt
\end{lstlisting}

\noindent Finally, we compile the CUDA operators as follows,
\begin{lstlisting}[frame=shadowbox]
cd ./models/ops
sh ./make.sh
# unit test (should see all checking is True)
python test.py
\end{lstlisting}

\paragraph{Running experiments.}

First, please download COCO 2017~\cite{Lin2014COCO}, and set up the dataset as in Deformable DETR~\cite{Zhu2021DeformableDETR}.

The following command runs the experiments:
\begin{lstlisting}[frame=shadowbox]
GPUS_PER_NODE=4 ./tools/run_dist_launch.sh 4 ./configs/r50_deformable_detr.sh
\end{lstlisting}

Settings can be changed in ``main.py''.

\input{supplementary/figures/fig_supp_visualization.tex}

%% file: supplementary/tables/supp_table_sota.tex
\begin{table*}
\small
\centering
\setlength{\tabcolsep}{3.25mm}{
\begin{tabular}{l|lcccccccc}
\toprule
Setting    & Method & Detection baseline & $AP$ & $AP_{50}$ & $AP_{75}$  & $AP_{S}$  & $AP_{M}$  & $AP_{L}$  \\ \midrule
     \multirow{6}{*}{$40${$+$}$40$}  & LwF \cite{LiH2018LwF} & GFLv1   & 17.2  & 25.4  & 18.6 & 7.9  & 18.4 & 24.3\\
     & RILOD \cite{Li2019RILOD} & GFLv1   & 29.9  & 45.0  & 32.0 & 15.8  & 33.0 & 40.5\\
     & SID \cite{Peng2021SID} & GFLv1   & 34.0  & 51.4  & 36.3 & 18.4  & 38.4 & 44.9\\
     & ERD \cite{Feng2022ElasticResponse} & GFLv1   & {36.9} & {54.5}  & {39.6}  & {21.3}  & {40.4}   & {47.5}\\ 
     & \cellcolor{mygray-bg}{CL-DETR \emph{w/o} ER} & \cellcolor{mygray-bg}{Deformable DETR} &  \cellcolor{mygray-bg}{{39.2}\tiny{$\pm$0.2}} & \cellcolor{mygray-bg}{{56.1}\tiny{$\pm$0.3}}  & \cellcolor{mygray-bg}{{42.6}\tiny{$\pm$0.4}} & \cellcolor{mygray-bg}{{21.0}\tiny{$\pm$0.3}}  & \cellcolor{mygray-bg}{{42.8}\tiny{$\pm$0.4}} & \cellcolor{mygray-bg}{{52.6}\tiny{$\pm$0.3}} \\
     \\[-12pt]
     & \cellcolor{mygray-bg}{CL-DETR} & \cellcolor{mygray-bg}{Deformable DETR} &  \cellcolor{mygray-bg}{\textbf{42.0}\tiny{$\pm$0.3}} & \cellcolor{mygray-bg}{\textbf{60.1}\tiny{$\pm$0.2}}  & \cellcolor{mygray-bg}{\textbf{45.9}\tiny{$\pm$0.3}} & \cellcolor{mygray-bg}{\textbf{24.0}\tiny{$\pm$0.3}}  & \cellcolor{mygray-bg}{\textbf{45.3}\tiny{$\pm$0.2}} & \cellcolor{mygray-bg}{\textbf{55.6}\tiny{$\pm$0.4}} \\
     \\[-12pt]
     \midrule
     \multirow{6}{*}{$70${$+$}$10$}  & LwF~\cite{LiH2018LwF} &  GFLv1   & 7.1  & 12.4  & 7.0 & 4.8  & 9.5 & 10.0\\
     & RILOD \cite{Li2019RILOD} & GFLv1   & 24.5 & 37.9  & 25.7 & 14.2  & 27.4 & 33.5\\
     & SID \cite{Peng2021SID} & GFLv1   & 32.8  & 49.0  & 35.0 & 17.1  & 36.9 & 44.5\\
     & ERD~\cite{Feng2022ElasticResponse} & GFLv1    & {34.9 } & {51.9}  & {37.4}  & {18.7}  & {38.8}   & {45.5}\\
     & \cellcolor{mygray-bg}{CL-DETR \emph{w/o} ER} & \cellcolor{mygray-bg}{Deformable DETR} &  \cellcolor{mygray-bg}{{35.8}\tiny{$\pm$0.3}} & \cellcolor{mygray-bg}{{53.5}\tiny{$\pm$0.2}}  & \cellcolor{mygray-bg}{{39.5}\tiny{$\pm$0.3}} & \cellcolor{mygray-bg}{{19.4}\tiny{$\pm$0.3}}  & \cellcolor{mygray-bg}{{41.5}\tiny{$\pm$0.3}} & \cellcolor{mygray-bg}{{46.1}\tiny{$\pm$0.4}} \\
     \\[-12pt]
     & \cellcolor{mygray-bg}{CL-DETR} & \cellcolor{mygray-bg}{Deformable DETR} &  \cellcolor{mygray-bg}{\textbf{40.4}\tiny{$\pm$0.2}} & \cellcolor{mygray-bg}{\textbf{58.0}\tiny{$\pm$0.3}}  & \cellcolor{mygray-bg}{\textbf{43.9}\tiny{$\pm$0.2}} & \cellcolor{mygray-bg}{\textbf{23.8}\tiny{$\pm$0.4}}  & \cellcolor{mygray-bg}{\textbf{43.6}\tiny{$\pm$0.3}} & \cellcolor{mygray-bg}{\textbf{53.5}\tiny{$\pm$0.3}} \\
     [-2pt]
\bottomrule 
\end{tabular}
}
\caption{\mycaptionsupp{Supplementary to Table~1 (main paper).} IOD results (\%) on COCO 2017 with the traditional IOD protocol~\cite{Feng2022ElasticResponse}. ``CL-DETR'' and ``CL-DETR \emph{w/o} ER'' are our methods. For ``CL-DETR \emph{w/o} ER'', we don't save any exemplars. For ``CL-DETR'',  the total memory budget for the exemplars is set as $10\%$ of the total dataset size. The results for the related methods~\cite{LiH2018LwF,Li2019RILOD,Peng2021SID,Feng2022ElasticResponse} are from ~\cite{Feng2022ElasticResponse}. 
In the $A+B$ setup, in the first phase, we observe a fraction $\frac{A}{A+B}$ of the training samples with $A$ categories annotated.
Then, in the second phase, we observe the remaining $\frac{B}{A+B}$ of the training samples, where $B$ new categories are annotated.
We test settings $A+B=40+40$ and $70+10$.
We run experiments for three different categories and data orders and report the average AP with $95\%$ confidence interval.
}%
\label{table_sota_supp}
\vspace{-0.1em}
\end{table*}

%% file: supplementary/tables/supp_table_lambda.tex
\newcommand{\tablespace}{\\[-12pt]}
\setlength{\tabcolsep}{0.9mm}{
\begin{table}[htp]
  \small
  \centering
  \begin{tabular}{lccc|ccccc}
  \toprule
   \tablespace
   Setting & KD  & Our KD & KD-oracle & ER & Our ER & ER-oracle\\
   \tablespace
   \midrule
   \tablespace
   AP & 24.5 & 33.9 & 36.1 & 33.3 & 36.1 & 36.5 \\
   \tablespace
  \bottomrule
\end{tabular}
  \caption{Ablation results (\%) for $\lambda$ on COCO 2017 in the $70+10$ setting.}%
  \label{table_rebuttal_1}
\end{table}
}

%% file: supplementary/tables/supp_table_separate.tex
\setlength{\tabcolsep}{1.7mm}{
\begin{table}[htp]
  \small
  \centering
  \begin{tabular}{lccc|ccccc}
  \toprule
   \tablespace
   Setting & $K$=$5$ & $K$=$10$ & $K$=$20$ & $p${$\geq$}$0.1$ & $p${$\geq$}$0.3$ & $p${$\geq$}$0.5$\\
   \tablespace
   \midrule
   \tablespace
   AP & 39.1 & 39.9 & 39.5 & 38.6 & 38.9 & 38.2 \\
   \tablespace
  \bottomrule
\end{tabular}
  \caption{Ablation results (\%) for different pseudo label selection strategies on a separate validation set (COCO 2017, $70+10$ setting).}%
  \label{table_rebuttal_4}
  \vspace{-0.15cm}
\end{table}
}

%% file: supplementary/tables/supp_table_curriculum.tex
\setlength{\tabcolsep}{2.5mm}{
\begin{table}[htp]
  \small
  \centering
  \begin{tabular}{lcccccccc}
  \toprule
   \tablespace
   Setting  & $p${$\geq$}$0.1$ & $p${$\geq$}$0.3$ & $p${$\geq$}$0.5$ & Curriculum for $p$\\
   \tablespace
   \midrule
   \tablespace
   AP  & 38.6 & 38.9 & 38.2 & 39.2\\
   \tablespace
  \bottomrule
\end{tabular}
  \caption{Ablation results (\%) for curriculum learning on COCO 2017 in the $70+10$ setting.}%
  \label{table_rebuttal_5}
\end{table}
}

%% file: supplementary/tables/supp_table_fine_grained.tex
\setlength{\tabcolsep}{0.1mm}{
\begin{table}[htp]
  \footnotesize
  \centering
  \begin{tabular}{lccccccccccccc}
  \toprule
  \\[-10pt]
   \multirow{1.5}{*}{{Method}}&\multicolumn{4}{c}{{All categories $\uparrow$}}&\multicolumn{4}{c}{{Old categories $\uparrow$}}&\multicolumn{4}{c}{{FPP $\downarrow$}}\\
   \\[-10pt]
   & $AP$ & $AP_{S}$ & $AP_{M}$ & $AP_{L}$& $AP$ & $AP_{S}$ & $AP_{M}$ & $AP_{L}$& $AP$ & $AP_{S}$ & $AP_{M}$ & $AP_{L}$\\
   \\[-10pt]
   \midrule
   \\[-10pt]
   LwF&24.5& 12.4& 28.2& 35.2&24.0& 12.3&27.7 & 34.4&19.3 & 13.5& 18.2& 23.1\\
   iCaRL&35.9&19.1&39.4&48.6&36.8&20.3&39.9&50.0&6.5&5.5&6.0&7.5\\
   Ours  & 40.1 & 23.2 & 43.2 & 52.1 & 41.8 & 24.5& 44.7 & 54.6 & 1.5 &1.3 &1.2 & 2.9\\
   \\[-10pt]
  \bottomrule
\end{tabular}
\caption{\mycaptionsupp{Supplementary to Table~2 (main paper).} More fine-grained ablation results (\%) for KD and ER, using Deformable DETR~\cite{Zhu2021DeformableDETR} on COCO 2017 in the $70+10$ setting.}%
  \label{table_rebuttal_6}
\end{table}
}

%% file: supplementary/tables/supp_table_ablation.tex
\begin{table*}
\small
\centering
\setlength{\tabcolsep}{0.90mm}{
\begin{tabular}{@{}c|c|cccccccccccccccccc@{}}
\toprule
  \multirow{3}{*}{{Row}} & \multirow{3}{*}{{Exemplar replay strategies}}& \multicolumn{4}{c}{\multirow{1.5}{*}{All categories $\uparrow$}}  && \multicolumn{4}{c}{\multirow{1.5}{*}{Old categories $\uparrow$}} && \multicolumn{4}{c}{\multirow{1.5}{*}{FPP $\downarrow$}}\\
   &  & \multirow{2.5}{*}{$AP$} & \multirow{2.5}{*}{$AP_{S}$}  & \multirow{2.5}{*}{$AP_{M}$}  & \multirow{2.5}{*}{$AP_{L}$} && \multirow{2.5}{*}{$AP$} & \multirow{2.5}{*}{$AP_{S}$}  & \multirow{2.5}{*}{$AP_{M}$}  & \multirow{2.5}{*}{$AP_{L}$} && \multirow{2.5}{*}{$AP$} & \multirow{2.5}{*}{$AP_{S}$}  & \multirow{2.5}{*}{$AP_{M}$}  & \multirow{2.5}{*}{$AP_{L}$} \\
  &    \\
    \midrule
    {1} &  Random &  37.9 & 20.8  & 40.9 & 50.4 && 39.0 & 21.6 & 41.7  & 52.3 && 4.3 & 4.2 & 4.2 & 5.2 \\
    {2} & Herding~\cite{Rebuffi2017iCaRL} &  38.1 & 22.5  & 41.0 & 49.3 && 39.0 & 23.2 & 41.6  & 50.4 && 4.3 & 2.6 & 4.3 & 7.1 \\
    {3} & Adaptive sampling~\cite{Liu2020MultiTask} &  38.5 & 22.7  & 41.4 & 49.9 && 39.4 & 23.5 & 42.1  & 51.2 && 3.9 & 2.3 & 3.8 & 6.3 \\
    \cellcolor{mygray-bg}{{4}} & \cellcolor{mygray-bg}{Distribution-preserving calibration (ours)} & \cellcolor{mygray-bg}{\textbf{40.1}} & \cellcolor{mygray-bg}{\textbf{23.2}}  & \cellcolor{mygray-bg}{\textbf{43.2}} & \cellcolor{mygray-bg}{\textbf{52.1}} &\cellcolor{mygray-bg}{}& \cellcolor{mygray-bg}{\textbf{41.8}} &\cellcolor{mygray-bg}{\textbf{24.5}} &\cellcolor{mygray-bg}{\textbf{44.7}} & \cellcolor{mygray-bg}{\textbf{54.6}} &\cellcolor{mygray-bg}{}& \cellcolor{mygray-bg}{\textbf{1.5}} & \cellcolor{mygray-bg}{\textbf{1.3}} & \cellcolor{mygray-bg}{\textbf{1.2}} & \cellcolor{mygray-bg}{\textbf{2.9}}\\
    [-2pt]
\bottomrule
\end{tabular}
}
\caption{\mycaptionsupp{Supplementary to Table~2 (main paper).} Ablation results (\%) for different exemplar replay strategies, using Deformable DETR~\cite{Zhu2021DeformableDETR} on COCO 2017 in the $70+10$ setting. ``Herding'' and ``adaptive sampling'' are from \cite{Rebuffi2017iCaRL} and \cite{Liu2020MultiTask}, respectively.}%
\label{table_supp_ablation}
\end{table*}

%% file: supplementary/figures/fig_supp_visualization.tex
\begin{figure*}[t!]
\centering
\includegraphics[width=1.02\textwidth]{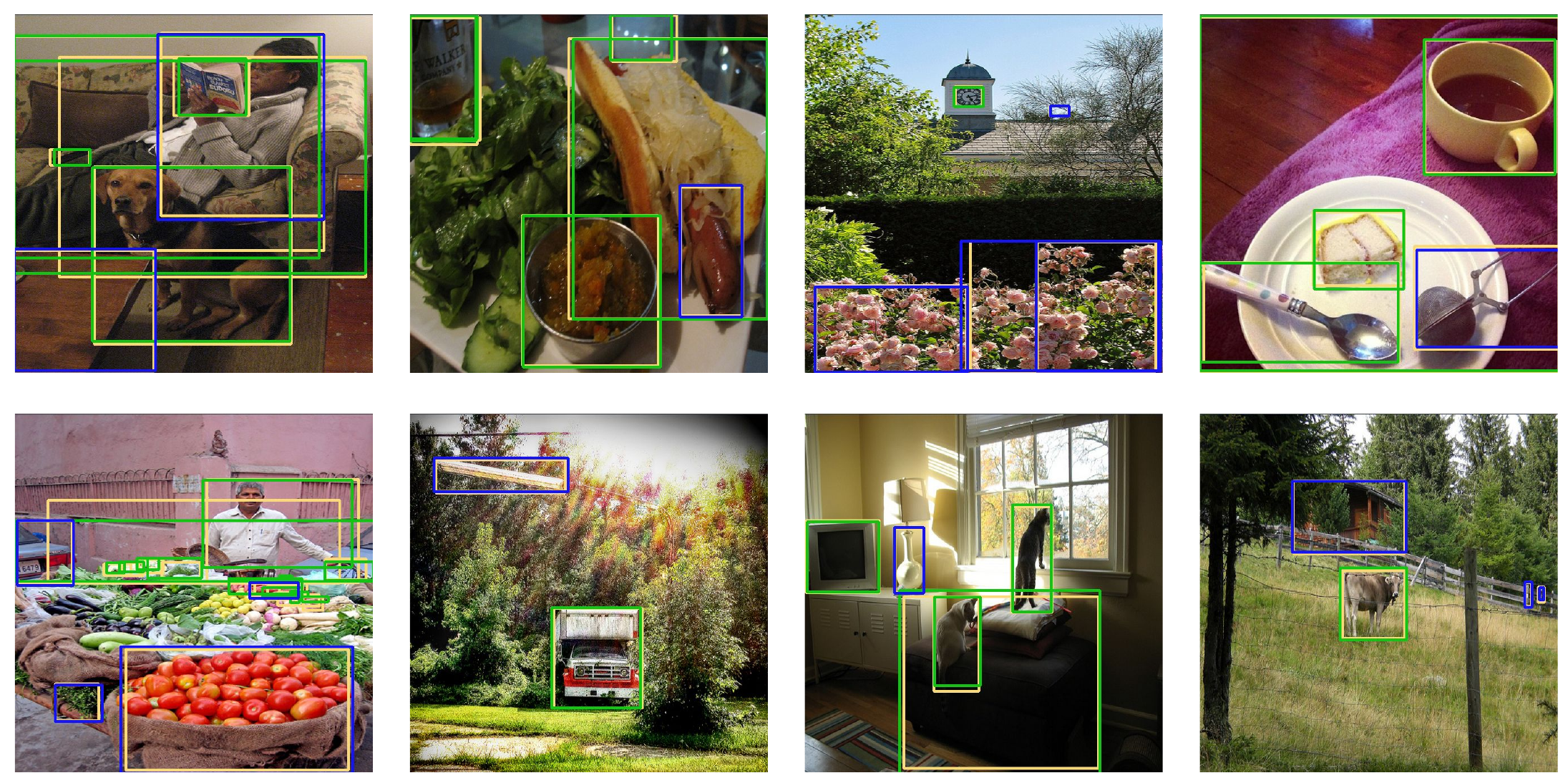}
\vspace{-0.2cm}
\caption{\mycaptionsupp{Supplementary to Section~4.2 (main paper).} Visualizations of the one-to-one matching between the merged bounding boxes and new model predictions (\textcolor[rgb]{0.97, 0.85, 0.43}{yellow}) on COCO 2017 using the $70+10$ setting. The merged bounding boxes include the old category pseudo 
 (\textcolor[rgb]{0.08, 0.08, 0.95}{blue}) and new category ground-truth (\textcolor[rgb]{0.08, 0.77, 0.08}{green}) bounding boxes. Our method ensures the old category pseudo and new category ground-truth bounding boxes are merged successfully.}%
\label{supp_fig_pairs}
\vspace{-0.1cm}
\end{figure*}